\documentclass[10pt,twocolumn,letterpaper]{article}

\usepackage{wacv}
\usepackage{times}
\usepackage{epsfig}
\usepackage{graphicx}
\usepackage{amsmath}
\usepackage{amssymb}
\usepackage{multimedia}
\usepackage{animate}
\usepackage{multirow}
\usepackage{tabularx}
\usepackage{enumerate}

\wacvfinalcopy 

\def\wacvPaperID{375} 

\begin{document}

\title{Single Image Deblurring and Camera Motion Estimation with Depth Map}

\author{Liyuan Pan $^{1,3}$, Yuchao Dai$^{2}$, and  Miaomiao Liu$^{1,3}$ \\
$^{1}$ Research School of Engineering, Australian National University, Canberra, Australia \\
$^{2}$ School of Electronics and Information, Northwestern Polytechnical University, Xi'an, China \\
$^{3}$ Australia Centre for Robotic Vision \\
\tt\small{\{liyuan.pan, miaomiao.liu\}}@anu.edu.au,
daiyuchao@nwpu.edu.cn
}

\maketitle

\begin{abstract}
Camera shake during exposure is a major problem in hand-held photography, as it causes image blur that destroys details in the captured images.~In the real world, such blur is mainly caused by both the camera motion and the complex scene structure.~While considerable existing approaches have been proposed based on various assumptions regarding the scene structure or the camera motion, few existing methods could handle the real 6 DoF camera motion.~In this paper, we propose to jointly estimate the 6 DoF camera motion and remove the non-uniform blur caused by camera motion by exploiting their underlying geometric relationships, with a single blurry image and its depth map (either direct depth measurements, or a learned depth map) as input.~We formulate our joint deblurring and 6 DoF camera motion estimation as an energy minimization problem which is solved in an alternative manner. Our model enables the recovery of the 6 DoF camera motion and the latent clean image, which could also achieve the goal of generating a sharp sequence from a single blurry image. Experiments on challenging real-world and synthetic datasets demonstrate that image blur from camera shake can be well addressed within our proposed framework.
\end{abstract}

\section{Introduction}

Image blurs are mainly caused by camera motions or motion of the objects in the scene during the long exposure time which is generally required under the low-light condition. It is a common problem for the hand-held photography and becomes increasingly important due to the popularity of the mobile devices such as smart phones in recent years. Blind image deblurring targets at recovering the latent clean images from the blurry ones. It has been an active research field in computer vision and image processing community~\cite{hyun2015generalized,sellent2016stereo,gong2017motion,Pan_2017_CVPR,Su_2017_CVPR}.

\begin{figure*}
\begin{center}
\begin{tabular}{cccc}
\hspace{-0.4cm}
\includegraphics[width=0.24\textwidth]{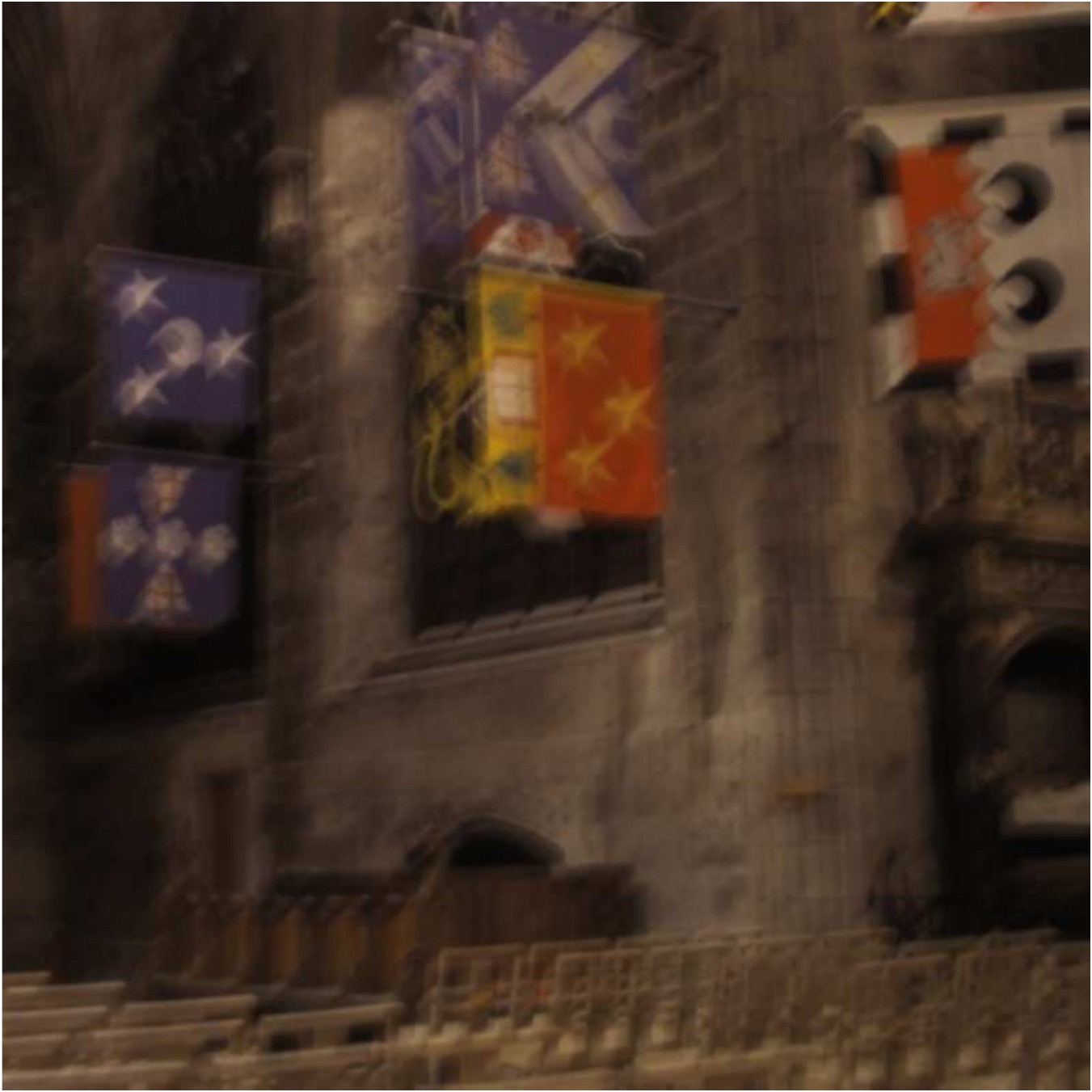}
&\begin{frame}{}
 \animategraphics[width=0.24\linewidth,autoplay,palindrome,loop]{12}{rebuttal/tmp-}{199}{201}
\end{frame}
&\includegraphics[width=0.24\textwidth]{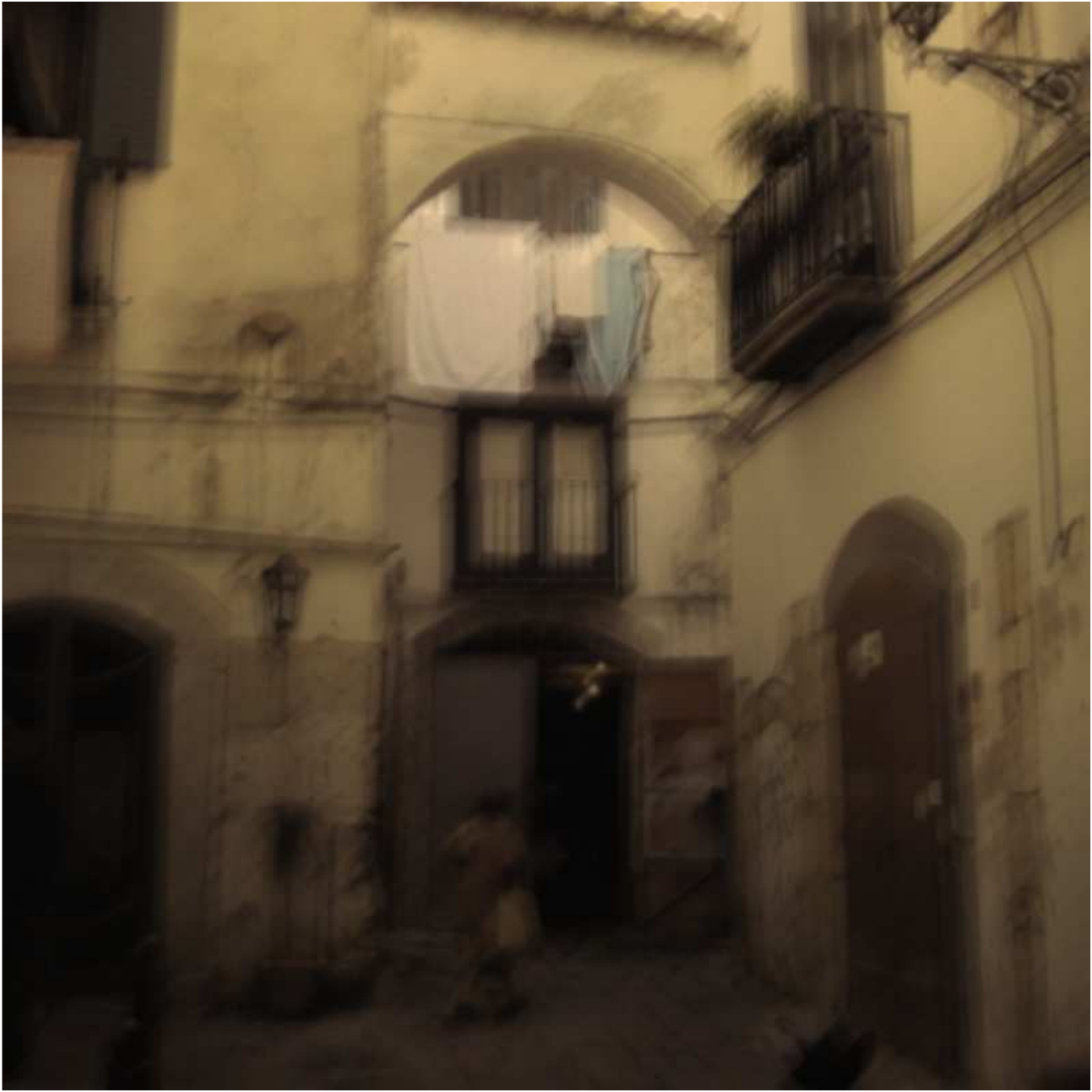}
&\begin{frame}{}
 \animategraphics[width=0.24\linewidth,autoplay,palindrome,loop]{12}{rebuttal/tmp-}{223}{225}
 \end{frame}\\
 \hspace{-0.4cm}
(a) Blurry Image & (b) Our Result & (c) Blurry Image & (d) Our Result  \\
\end{tabular}
\end{center}
\caption{\label{fig:fig1} (a), (c) are the input blurry images from \cite{kohler2012recording} dataset. (b), (d) are our deblurring results. We first use the input blurry image  to learn a depth map by using \cite{godard2017unsupervised}. Then, we jointly estimate camera motion and deblur the image with the learned depth map. With the depth map and the 6 Dof camera pose, we can project the recovered image to a sharp image sequence. We display one image of our deblurring sequence (during the exposure time). (Best view in Adobe Reader)
}
\end{figure*}

Blind image deblurring is a very challenging task since it is highly under-constrained as multiple pairs of blur kernels and latent images can generate the same blurry image. A single blur kernel cannot model the complex blurs in real-world scenarios. Existing methods have exploited various constraints to model the characteristics of blur and utilize different natural image priors to regularize the solution space \cite{Comparative_Study_Blind_Deblurring:CVPR-2016}. However, these assumptions, such as uniform blur \cite{xu2013unnatural}, non-uniform blur from multiple homography \cite{hu2014joint,pan2016soft}, with moving objects~\cite{pan2016soft}, constant depth \cite{gupta2010single,xu2012depth}, in-plane rotation \cite{sun2015learning}, forward motion \cite{Zheng_2013_ICCV} may not be satisfied and applicable in practice. 

In this paper, we focus on estimating and removing the spatially-varying motion blur caused by camera shake during the exposure time and propose to achieve blind image deblurring by explicitly exploiting the 6 DoF (degrees-of-freedom) camera motion (see Fig.~\ref{fig:fig1} for an example). In our formulation, the observed blurry image is formed by a composition of both the 6 DoF camera motion and the 3D scene structure, which enables us to capture the real blurry image generation process especially due to camera shake.

In order to handle the real world spatially-variant blur, we make the following assumptions regarding the scene structure and the camera motion:
\begin{enumerate}[1)]
\item \textbf{Availability of depth map of the scene.}
As more and more consumer cameras are now equipped with depth sensors such as iPhone X, the availability of depth map becomes a rather reasonable and realistic assumption.
Furthermore, the advent of deep learning also enables the estimation of a dense depth map from a single color image (monocular depth estimation) \cite{ranftl2016dense,liwicki2016coarse,li2017monocular}.
\item \textbf{Small camera motion.} Due to the short exposure time and the high sampling rate of modern video cameras, the~\emph{camera~shake process} can be modeled as the camera essentially undergoes a motion with small rotation angle and linear translation.~We thus adopt the small angle approximation of rotation~\cite{Accidental-Motion:CVPR-2014}.
\end{enumerate}

The above assumptions naturally lead to a few legitimate queries:
\begin{enumerate}[1)]
\item \textbf{Why the 6 DoF camera motion is needed?} Recently, several deep learning based approaches \cite{Jin_2018_CVPR,purohit2018bringing} could restore a video from a single blur image. However, the restored video sequence is not guaranteed to respect the 3D geometry of the scene as well as the camera motion. Instead, we target at recovering the 6 DoF camera motion which allows the recovery of a sharp video sequence from a single blurry image as well as the capacity of novel view synthesis for high frame rate video sequences. In Fig.~\ref{fig:fig1}, we illustrate the recovered video sequence from a single blurry image, which clearly demonstrates the benefit of our camera motion model.
\vspace{-4mm}
\item \textbf{Why the small camera motion model is useful?} For small rotation model, the simplified rotation matrix is robust to noise as the second-order Taylor expansion of the rotation matrix has been ignored. The small motion model has been proven to be the key in estimating the camera poses and the 3D structure in the context of 3D reconstruction from accidental motion \cite{Im_2015_ICCV}. More complex camera trajectories could be exploited with the cost of increasing computational complexity.
\end{enumerate}

\vspace{-1.5mm}
Building upon the above assumptions regarding the camera motion and the scene structure, we formulate blind image deblurring as the task of joint latent clean image recovery and 6 DoF camera motion estimation \footnote{The most similar work to ours seems to be Park and Lee \cite{Joint-Pose-Depth-Deblurring-Super:ICCV-2017}, which solves for camera pose, scene depth, deblurring image and super-resolution under a unified framework from a \textbf{image sequence}. Different from \cite{Joint-Pose-Depth-Deblurring-Super:ICCV-2017}, our method takes \textbf{a single blurry image and a depth map} as input to achieve camera motion estimation and image deblurring}. Our unified framework naturally relates camera motion estimation and image deblurring, where the solution of one sub-task benefits the solution of the other sub-task. Specifically, we present an energy minimization based framework which involves both a unary term in explaining the observed blurry image and regularization terms on the camera motion and the desired latent clean image. To speed up the implementation and provide effective optimization, we apply a coarse-to-fine strategy to the energy minimization, where in each level we perform camera motion estimation and image deblurring in an alternative manner.

\vspace{1mm}

Our main contributions can be summarized as:

\begin{itemize}
\item We propose to jointly estimate the 6 DoF camera motion and deblur the image from a blurry image while giving its depth map (the depth information is from depth measurements or learned from the color image);
\vspace{-2mm}
\item We propose to use the small motion camera model which not only simplifies the motion estimation problem but also leads to an efficient solution;
\item Extensive experiments on both synthetic and real images prove the effectiveness of our method especially its robustness against noisy depth maps.
\vspace{-2mm}
\end{itemize}
\section{Related Work}

Recently, significant progress has been made in blind image deblurring. As there is a rich family of image deblurring methods, here we confine ourself to the most related ones. Blind image deblurring methods could be roughly categorized into two groups: monocular methods (image and video) and multi-view methods. Besides, we will also briefly cover deep learning based deblurring approaches.

\vspace{1mm}
\noindent{\bf{Monocular image deblurring.}}
Blind image deblurring is highly ill-posed, therefore various constraints on the blur kernels or the latent images have been proposed to regularize the solution space, which include the gradient based regularizers such as total variation \cite{Pan_2017_CVPR}, Gaussian scale mixture \cite{fergus2006removing}, $l_1 \backslash l_2$ norm \cite{krishnan2011blind}, and the ${l}_0$-norm regularizer \cite{xu2013unnatural}. Besides, non-gradient-based priors such as the color line based prior \cite{lai2015blur}, and the extreme channel (dark/bright channel) prior \cite{pan2016blind,pan2017deblurring,yan2017image} have also been explored. 
The fact that blur caused by camera shake in images are usually non-uniform motivates a series of work in modeling the spatially-variant blur. Whyte \etal \cite{whyte2012non} approximated the blur kernels by discretization in the space of 3D camera rotations. Gupta \etal \cite{gupta2010single} used a motion density function to represent the camera motion trajectory for the non-uniform deblurring, which requires the constant depth or fronto-parallel scene assumption.
Hirsch \etal \cite{hirsch2011fast} assumed that blur is locally invariant and proposed a fast non-uniform framework based on efficient filter flow. Zheng \etal \cite{Zheng_2013_ICCV} considered only discretized 3D translations.
Hu \etal \cite{hu2014joint} proposed to jointly estimate the depth layering and remove non-uniform blur caused by in-plane motion from a single blurry image, which, however, requires user input for depth layers partition and known depth values a prior.  Pan \etal \cite{pan2016soft} proposed to jointly estimate object segmentation and camera motion by incorporating soft segmentation, but requires user input. In practical settings, it is still challenging to remove strongly non-uniform motion blur in complex scenes.

\vspace{1mm}
\noindent{\bf{Video deblurring.}} Single image based deblurring has been extended to video sequence to better remove blurs in dynamic scenes\cite{cho2012video,kim2014segmentation,hyun2015generalized,Pan_2017_CVPR}.
Wulff and Black \cite{wulff2014modeling} proposed a layered model to estimate both foreground motion and background motion. However, these motions are restricted to affine models, and it is difficult to be extended to multi-layer scenes due to the requirement of depth ordering of the layers.
Kim \etal \cite{hyun2015generalized} proposed to simultaneously estimate optical flow and tackle the case of general blur by minimizing a single non-convex energy function.
As depth can significantly simplify the deblurring problem, multi-view deblurring methods have been proposed to leverage the depth information. Xu \etal~\cite{xu2012depth} inferred depth from two blurry images captured by a stereo camera and proposed a hierarchical estimation framework to remove motion blur caused by in-plane translation. Sellent \etal \cite{sellent2016stereo} proposed a stereo video deblurring technique, where 3D scene flow is estimated from the blur images using a piecewise rigid scene representation. Pan \etal \cite{Pan_2017_CVPR} proposed a single framework to jointly estimate the scene flow and deblur the images. Lee \etal \cite{Lee_2018_ECCV} proposed to estimate all blur model variables jointly, including latent sub-aperture image, camera motion, and scene depth from the blurred 4D light field.

\vspace{1mm}
\noindent{\bf Deep learning based image deblurring.}
Recently, the success of deep learning in high-level vision tasks have also been extended to low-level vision tasks such as image deblurring\cite{vasu2017local,kim2017online,Su_2017_CVPR,Nah_2017_CVPR,Tao_2018_CVPR}. Sun \etal \cite{sun2015learning} proposed a convolutional neural network (CNN) to estimate locally linear blur kernels. Gong \etal \cite{gong2017motion} learned optical flow field from a single blurry image directly through a fully-convolutional deep neural network and recovered the clean image from the learned optical flow. Jin \etal \cite{Jin_2018_CVPR} extracted a video sequence from a single motion-blurred image by introducing loss functions invariant to the temporal order.
Li \etal \cite{li2018learning} used a learned image prior to distinguish whether an image is sharp or not and embedded the learned prior into the MAP framework. Tao \etal \cite{Tao_2018_CVPR} proposed a light and compact network, SRN-DeblurNet, to deblur the image. With the supervised learning nature of these deep learning based deblurring methods, the success strongly depends on the statistical consistency between the training datasets and the testing datasets, which could hinder the generalization ability for real world applications.

\begin{figure*}[!ht]
\begin{tabular}{c}
\includegraphics[width=0.90\textwidth]{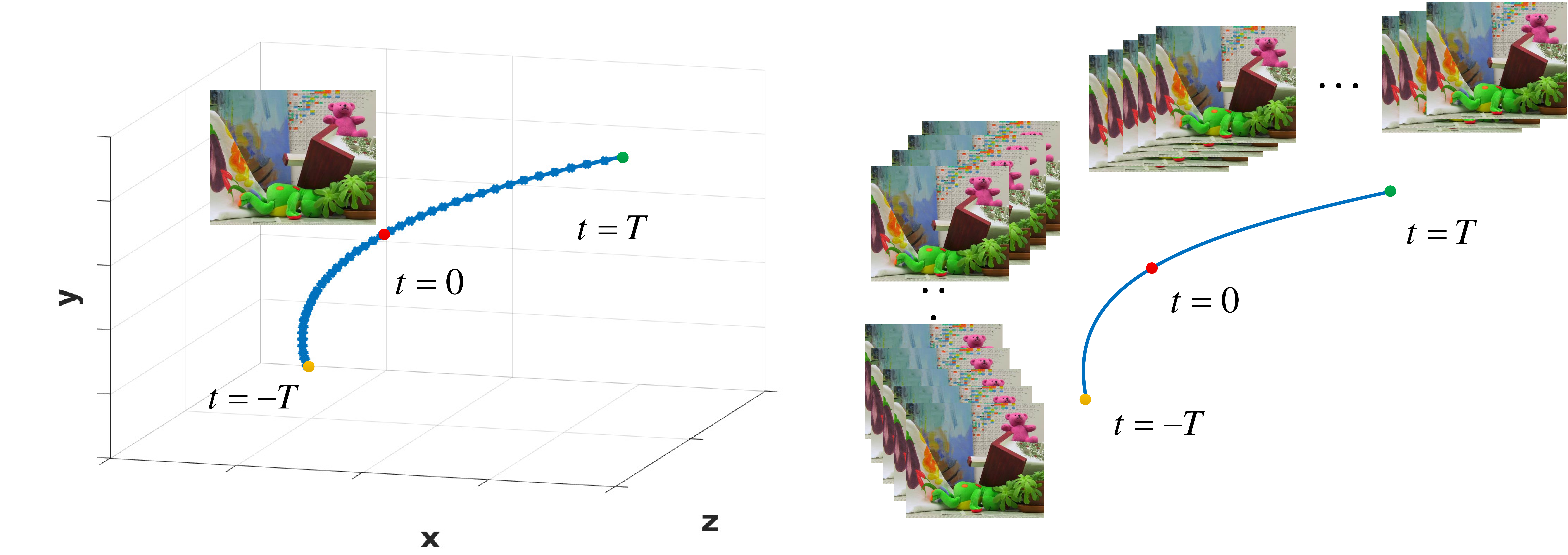}\\
\end{tabular}
\caption{\label{fig:pipeline} Example of our blur model. We approximate the blurry image by averaging the images sequence during the exposure time $2T$, where the spatially-variant blur kernel induced by the 6 DoF camera motion. (Best viewed on screen).}
\end{figure*}

\section{A Unified Spatially-varying Camera Shake Blur Model}

In this section, we develop a unified spatially-varying camera shake blur model, which explicitly relates the 6 DoF camera motion (including in/out-of-plane rotation and translation), and the latent clean image. In particular, we formulate our problem as a joint estimation of 6 DoF camera motion and image deblurring for depth-varying scenery.
\subsection{Blur Model}
\label{sec:blurmodel}

Given a single blurry image $\mathbf{B}$ and its corresponding depth map $\mathbf{D}$ (either from depth sensors or learned through a deep neural network), our goal is to find a clean (latent) image $\mathbf{L}$ and its corresponding camera motion during image capture. The blurry image can be modeled as a convolution of the latent image with a spatially-varying blur kernel $\mathbf{k}_{\bf x}$,
\begin{equation}
\mathbf{B}({\bf x}) =( \mathbf{k}_{\bf x}\otimes\mathbf{L})({\bf x})+z,
\end{equation}
where ${\bf k_x}$ denotes the blur kernel at pixel location ${\bf x}\in {\mathbb{R}^2}$,
$\otimes$ is the convolution operator, $z \sim \mathcal{N}(0,\sigma^2)$ is defined as the Gaussian noise. Note that this problem is highly under-determined since multiple pairs of $\mathbf{L}$ and $\mathbf{k}_{\bf x}$ could lead to the same blurry image. We therefore make assumptions on the generation process of the blur image that, for complex dynamic settings such as outdoor traffic scenes, the spatially-varying blur kernels are determined by the 6 DoF camera motion and the scene structure.

The blurry image is generally modeled as the integration of the images during the exposure time $2T$. In our model, we will explicitly model the blurry image generation process with respect to the 6 DoF camera motion. Given the depth map $\mathbf{D}$ corresponding to the latent image $\mathbf{L}$ and the camera motion $\mathbf{p_t}$, the image at time $t$ is defined as $w(\mathbf{p}_t,\mathbf{D},\mathbf{L})$. $w(\cdot)$ is referred as the warping function which is defined by the back-projection of the latent image to 3D points based on the depth $\mathbf{D}$ followed by a forward projection to image frame at time $t$ based on $\mathbf{p}_t$.
The blurry image is therefore generated as
{\small
\begin{equation} \label{eq:convBlurKernel1}
\mathbf{B} = \lambda_T\int^{T}_{t=-T} w(\mathbf{p}_t,\mathbf{D},\mathbf{L}) dt + z,
\end{equation}
}
where $\lambda_T = \frac{1}{2T}$.~In general, we handle the problem in discrete space with
{\small
\begin{eqnarray} \label{eq:convBlurKernel}
\mathbf{B} &=&\lambda_N \sum_{n =-N}^N w(\mathbf{p}_{t},\mathbf{D},\mathbf{L})+z,\\
&=&\mathbf{A}_{\mathbf{p}}(\mathbf{L}) + z,\nonumber
\end{eqnarray}
}
where sample frequency $\lambda_N = \frac{1}{2N+1}$, $N$ is the sample number, and $n$ is the sample index.

\subsection{Camera Motion Model}

We further assume that the camera performs uniform out-of-plane rotation and translation. Let $\mathbf{p} = (\theta_x, \theta_y, \theta_z, v_x, v_y, v_z)^T$ represent the absolute motion during the exposure time $2T$.
The camera motion at time $t$, is then defined as $\mathbf{p}_{t} = (t/2T)*\mathbf{p}$. Let $ {\mathbf{\theta}} = (\theta_x, \theta_y, \theta_z)^T$ be the rotation parameters (Rodrigues' rotation formula \cite{belongie1999rodrigues}), and $\mathbf{v} = (v_x, v_y, v_z)^T$ be the translation vector. Since the camera exposure time is usually very short (several milliseconds), we assume that the camera performs small rotation motion, thus the rotation matrix can be approximated as
{\small
\begin{equation}
\mathbf{R} = \mathbf{I} + [\mathbf{\theta}]_{\times} = \begin{bmatrix}
1        & -\theta_z & \theta_y \\
\theta_z & 1         & -\theta_x\\
-\theta_y & \theta_x  & 1   \nonumber
\end{bmatrix},
\end{equation}
}
where $[\cdot]_{\times}$ denotes the cross-product operator, and $\mathbf{I}$ is the identity matrix. The small rotation motion assumption results in a first-order approximation of the rotation matrix.

Based on the above blur model and small motion model, we define our energy functions for deblurring and camera motion estimation in the following sections.

\subsection{Energy Formulation}
Our energy function is defined on the latent clean image and the 6 DoF camera motion. We formulate our problem in a unified framework to jointly estimate the camera motion and deblur the image. Our energy function is defined as
{\small
\begin{equation}
\label{eq:energy}
E = E_{\mathrm{blur}}(\mathbf{L,p})  + E_{\mathrm{reg}}(\mathbf{L,p}),
\end{equation}
}
which consists of a data term for deblurring, a regularization term enforcing the smoothness in camera motion, induced optical flow and the latent clean image.
The energy function terms are further discussed in the following sections.

\subsubsection{Data Term for Deblurring.}
\label{sec:dataTerm}
Our data term for deblurring involves two terms, which is defined as
\begin{equation}
\small
E_{\mathrm{blur}}(\mathbf{L,p}) =  \left\|  \mathbf{A_p(L)} - \mathbf{B}\right\|_{\mathrm{F}}^2 + \left\| \nabla  \mathbf{A_p(L)}- \nabla \mathbf{B}\right\|_{\mathrm{F}}^2.
\end{equation}
The first term encodes the fact that the estimated blur image from spatially-varying blur kernel should be similar to the observed blurry image. The second term encourages the intensity changes (gradient) in the estimated blurry image should be close to that of the observed blurry image.

\subsubsection{Regularization Terms.}
Our regularization terms explore the small motion constraints on the camera motion model, spatial smoothness constraints on the latent image and optical flow induced by the camera motion. The first one is to avoid the trivial solution of $\mathbf{p = 0}$. The second one is to enforce the optical flow generated from the camera motion and the depth map to be smooth across the image and respect the image and depth discontinuities. The third term is to suppress the noise in the latent image and penalize the spatial fluctuations.
~To this end, our potential function is defined as
{\small
\begin{equation}
E_{\mathrm{reg}}(\mathbf{p},\mathbf{L}) = \mu_1 \left\| \mathbf{p} \right\|_2^{2}  + S(\mathbf{p}) + \mu_4 \left\|\nabla\mathbf{L}_{(i,j)}\right\|_1,
\end{equation}
}
where $S(\mathbf{p})$ is defined as
\begin{equation}
\small
\begin{aligned}
S(\mathbf{p}) &= \mathcal{E}(B,D) \left\| \nabla F(\mathbf{p})_{(i,j)}\right\|_2^2\nonumber,\\
\mathcal{E}(B,D) &= \sum_{i,j\in \mathrm{\Omega}}\mu_2\ e^{\left(-\frac{\left\| \nabla \mathbf{B}_{(i,j)}\right\|_2^2} {\sigma^2_\mathrm{B}}\right) }+ \mu_3\ e^{\left(-\frac{\left\| \nabla \mathbf{D}_{(i,j)}\right\|_2^2}{ \sigma^2_\mathrm{D}}\right)}.\\
\end{aligned}
\end{equation}


$\mathrm{\Omega}$ denotes the image region. $\mu_{\{1,2,3,4\}}$ are weight parameters with $\mu_1  < 0$. $\sigma_{\{\mathrm{B,D}\}}$ are parameters for balancing the influence of the image and depth discontinuity on the spatial smoothness constraints. $F(\mathbf{p})$ denotes the optical flow field induced by camera motion $\mathbf{p}$ and depth map $\mathbf{D}$, which is obtained by forward projection of the 3D points corresponding to $\mathbf{t=0}$ to the camera motion $\mathbf{p}$.

\section{Solution}\label{sec:optimization}
The optimization of our energy function defined in Eq.~(\ref{eq:energy}), is to solve two different sets of variables, which are the camera motion $\mathbf{p}$ and the latent image $\mathbf{L}$, respectively. In order to solve the variables more efficiently, we perform the optimization alternatively through the following steps,
\begin{itemize}
\item Fix the latent image $\mathbf{L}$, solve for the camera motion $\mathbf{p}$ by optimizing Eq.~(\ref{eq:sceneFlowEnergy}) (See Section~\ref{sec:sceneflow}).
\item Fix the motion parameters $\mathbf{p}$, solve for the latent image $\mathbf{L}$ by optimizing Eq.~(\ref{eq:latentImageEnergy}) (See Section~\ref{sec:deblurring}).
\end{itemize}
In the following sections, we describe the details for each optimization step.

\subsection{Camera motion estimation}\label{sec:sceneflow}
We fix the latent image, namely $\mathbf{L} = \tilde{\mathbf{L}}$, then Eq.~(\ref{eq:energy}) reduces to
{\small
\begin{equation}\label{eq:sceneFlowEnergy}
\begin{aligned}
\min_{\mathbf{p}} \left\|  \mathbf{A_p(\tilde{L})} - \mathbf{B}\right\|_\mathrm{F}^2 &+  \left\| \nabla \mathbf{A_p(\tilde{L})} - \nabla \mathbf{B}\right\|_\mathrm{F}^2 \\
&+ \mu_1 \left\| \mathbf{p} \right\|_2^{2}+ S(\mathbf{p}).
%
\end{aligned}
\end{equation}
}
This is a non-linear and non-convex optimization problem. Fortunately, the solution space (6 DoF camera motion) is very small. We solve the problem by a nonlinear least-squares method \cite{more1978levenberg} to find the solution.

\subsection{Image deblurring}\label{sec:deblurring}
Given the 6 DoF camera motion parameters, namely $\tilde{\mathbf{p}}$, the blur image is derived based on Eq.~(\ref{eq:convBlurKernel}). The objective function in Eq.~(\ref{eq:energy}) becomes convex with respect to $\mathbf{L}$ and is expressed as
{\small
\begin{equation}\label{eq:latentImageEnergy}
\small
\min_{\mathbf{L}} \left\| \mathbf{A_{\tilde{p}}(L)} - \mathbf{B}\right\|_\mathrm{F}^2 + \left\| \nabla \mathbf{A_{\tilde{p}}(L)}  - \nabla \mathbf{B}\right\|_\mathrm{F}^2
+ \mu_4 \left\|\mathbf{L}\right\|_\mathrm{TV}.
\end{equation}
}
In order to obtain the latent clean image $\mathbf{L}$, we adopt the conventional primal-dual optimization method \cite{chambolle2011first} and derive the updating scheme as follows
{\small
\begin{equation}
\small
\left\{
\begin{gathered}
\begin{aligned}
 \mathbf{q}^{r+1}&=\frac{\mathbf{q}^{r}+\gamma \nabla \mathbf{L}^r}{\mathbf{max}(1,|(\mathbf{q}^{r}+\gamma\nabla \mathbf{L}^r)|)},\\
 \mathbf{L}^{r+1} &=\arg \min_{\mathbf{L}} \left\| \mathbf{A_{\tilde{p}}(L)}  - \mathbf{B}\right\|^2 +  \left\| \nabla \mathbf{A_{\tilde{p}}(L)}  - \nabla \mathbf{B}\right\|^2 \\
& + \frac{\left\| \mathbf{L}^{r+1}-(\mathbf{L}^{r}-\eta (\mu_4 \nabla {\bf q}^{r+1}) \right\|^2}{2\eta},
\end{aligned}
\end{gathered}\right.
\end{equation}
}
where $r$ is the iteration number, $q^{r}$ denotes the dual variable, $\eta = 10$ and $\gamma = 0.005$ are update step parameters. More details are referred to \cite{chambolle2011first}.

To further speed up the alternative optimization, we propose to apply a coarse-to-fine strategy to the energy minimization. Specifically, we perform camera motion estimation and image deblurring in an alternative manner in each level. The results from the coarse levels can be used as initialization for the following fine levels. 







\vspace{-0.2cm}
\section{Experiments}\label{sec:experiments}

\vspace{-0.1cm}
\subsection{Experimental Setup}


\begin{table}[t]
\centering
\caption{Comparison of flow error and deblurring results on different datasets (Middlebury, KITTI and TUM).}
\label{all}{\small
\begin{tabular}{c|c|c|c|c|c}
\hline
\multicolumn{2}{c|}{}                                                             & Pan \cite{pan2017deblurring}   & Yan \cite{yan2017image}   & kim \cite{hyun2015generalized}   & Our    \\ \hline
\multirow{2}{*}{\begin{tabular}[c]{@{}c@{}}PSNR\\ (dB)\end{tabular}} &   \begin{tabular}[c]{@{}c@{}}Middle-\\ burry\end{tabular} & 25.44  & 24.98  & -     & {\bf 26.16}  \\ \cline{2-6}
                                                                     & KITTI       & 22.78  & 23.28  & -     & {\bf 26.21}  \\ \hline
\multirow{2}{*}{SSIM}                                                &  \begin{tabular}[c]{@{}c@{}}Middle-\\ burry\end{tabular} & 0.7962 & 0.7822 & -     & {\bf 0.8357} \\ \cline{2-6}
                                                                     & KITTI       & 0.7615 & 0.7715 & -     & {\bf 0.8289} \\ \hline
{\small\begin{tabular}[c]{@{}c@{}}Flow\\ Error\end{tabular}}            & TUM         & -      & -      & 31.95 & {\bf 27.57}  \\ \hline
\end{tabular}}
\end{table}


\begin{figure*}[!ht]
\begin{center}
\resizebox{\textwidth}{!}{
\begin{tabular}{cccc}
\hspace{-0.35 cm}
\includegraphics[width=0.23\textwidth]{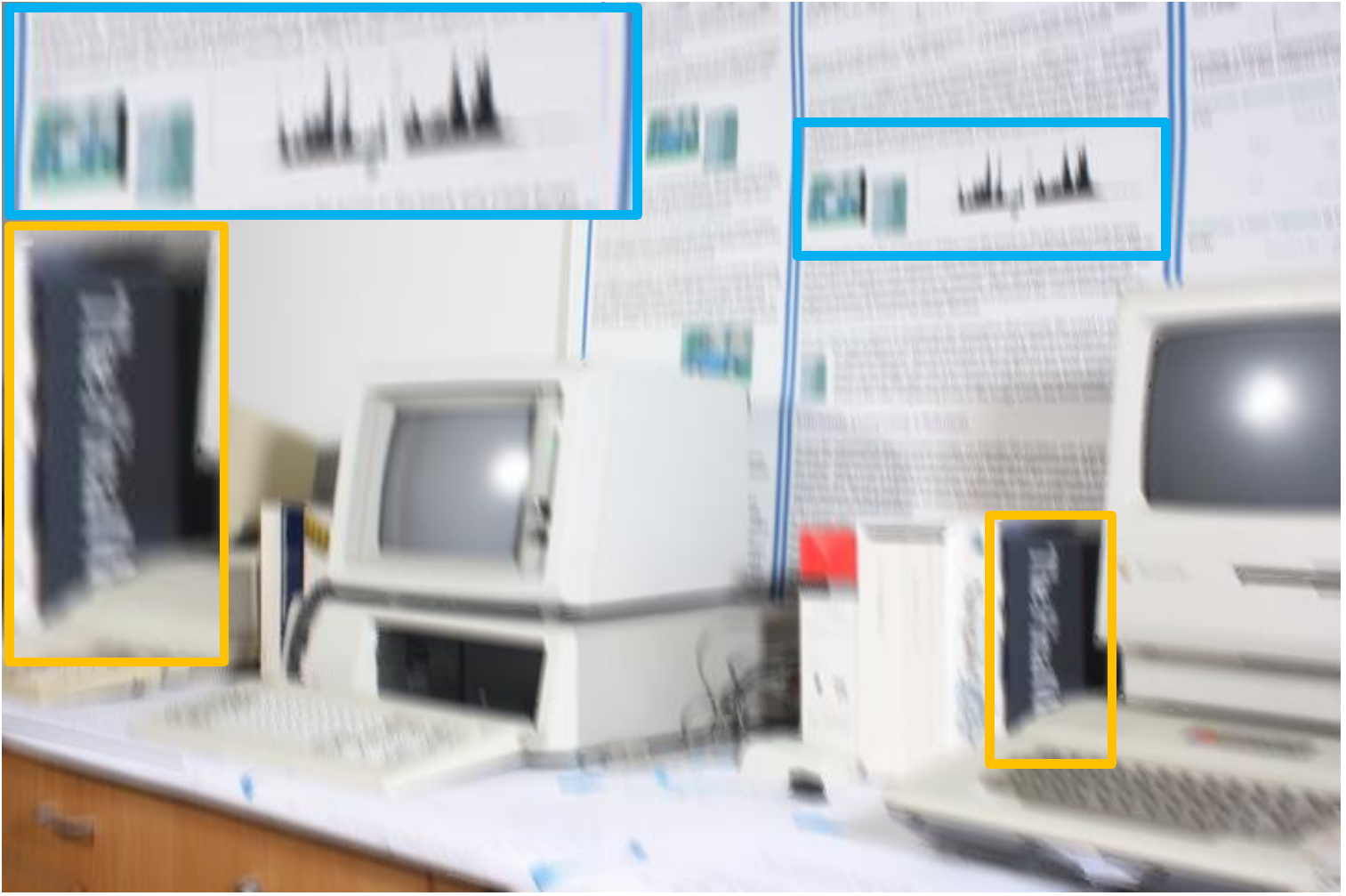}
&\includegraphics[width=0.23\textwidth]{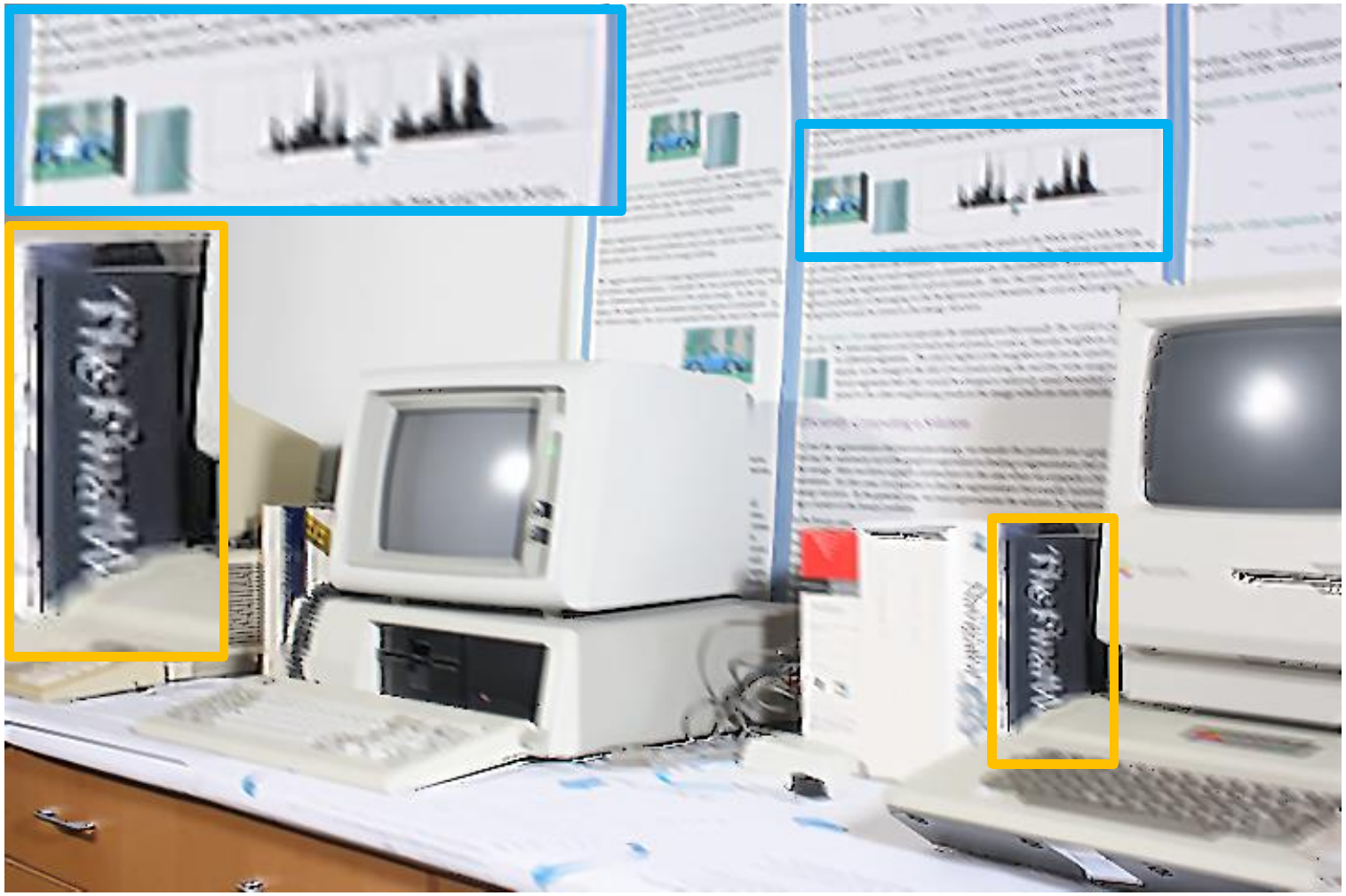}
&\includegraphics[width=0.23\textwidth]{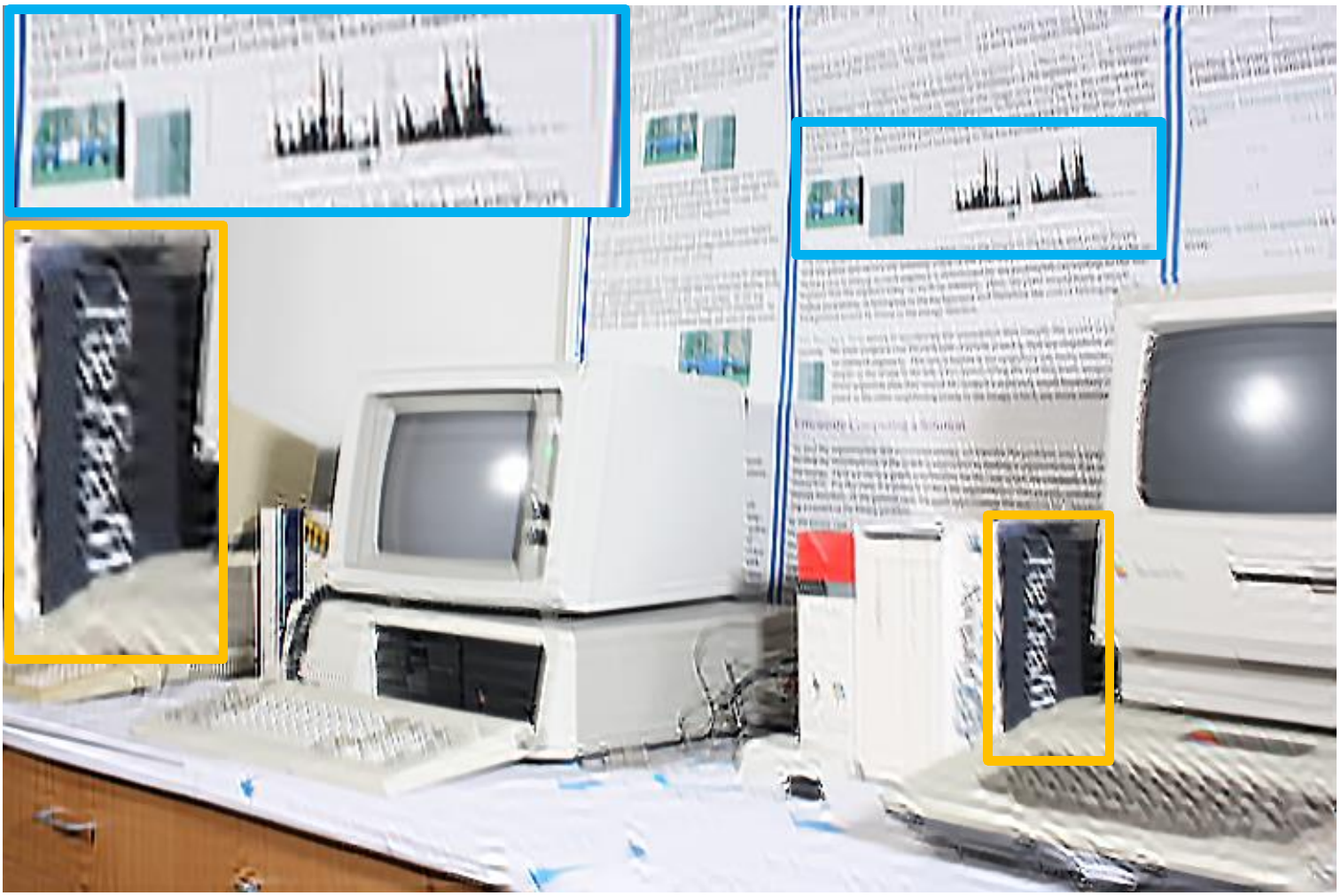}
&\includegraphics[width=0.23\textwidth]{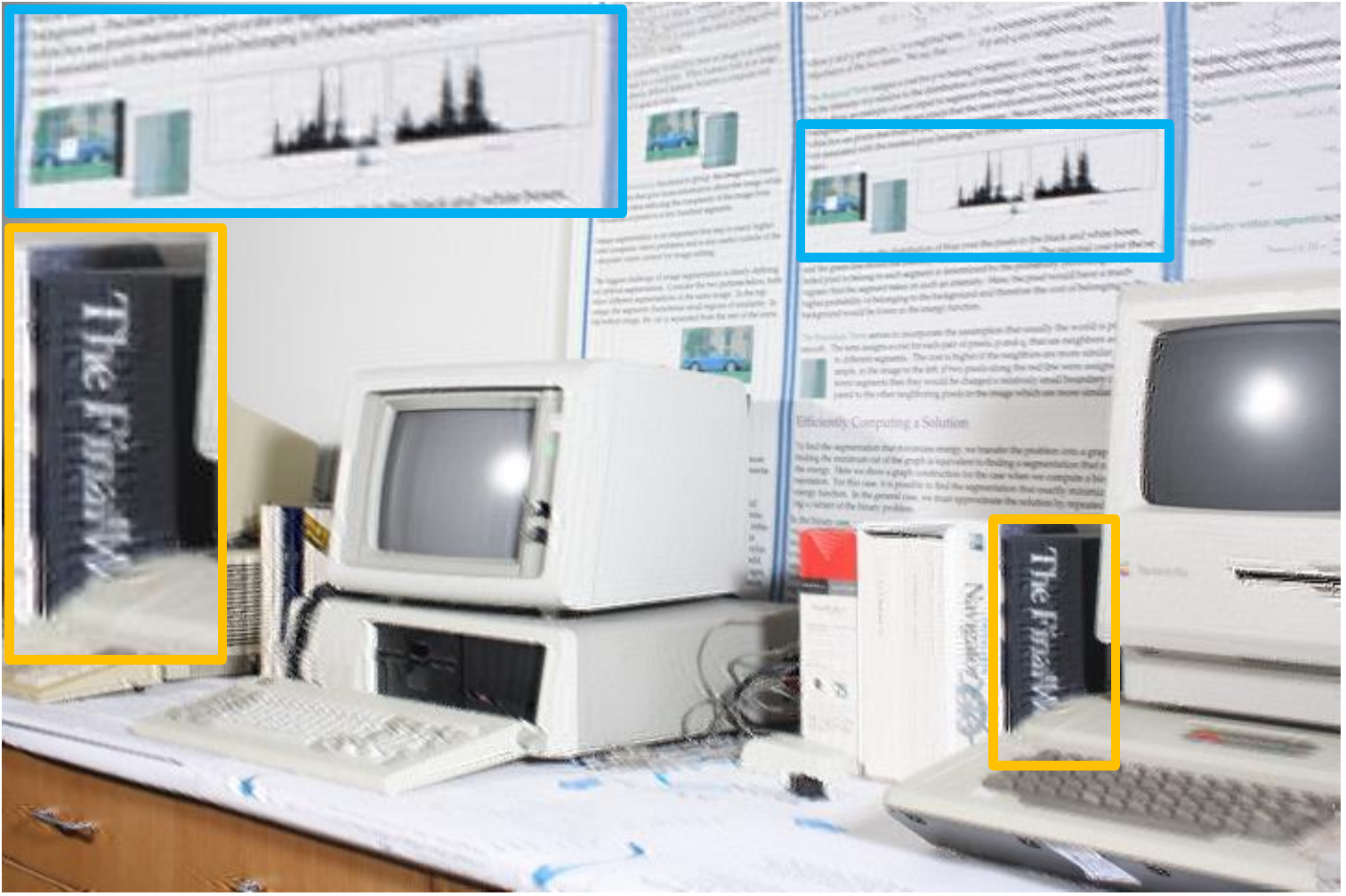}\\
\hspace{-0.35 cm}
\includegraphics[width=0.23\textwidth]{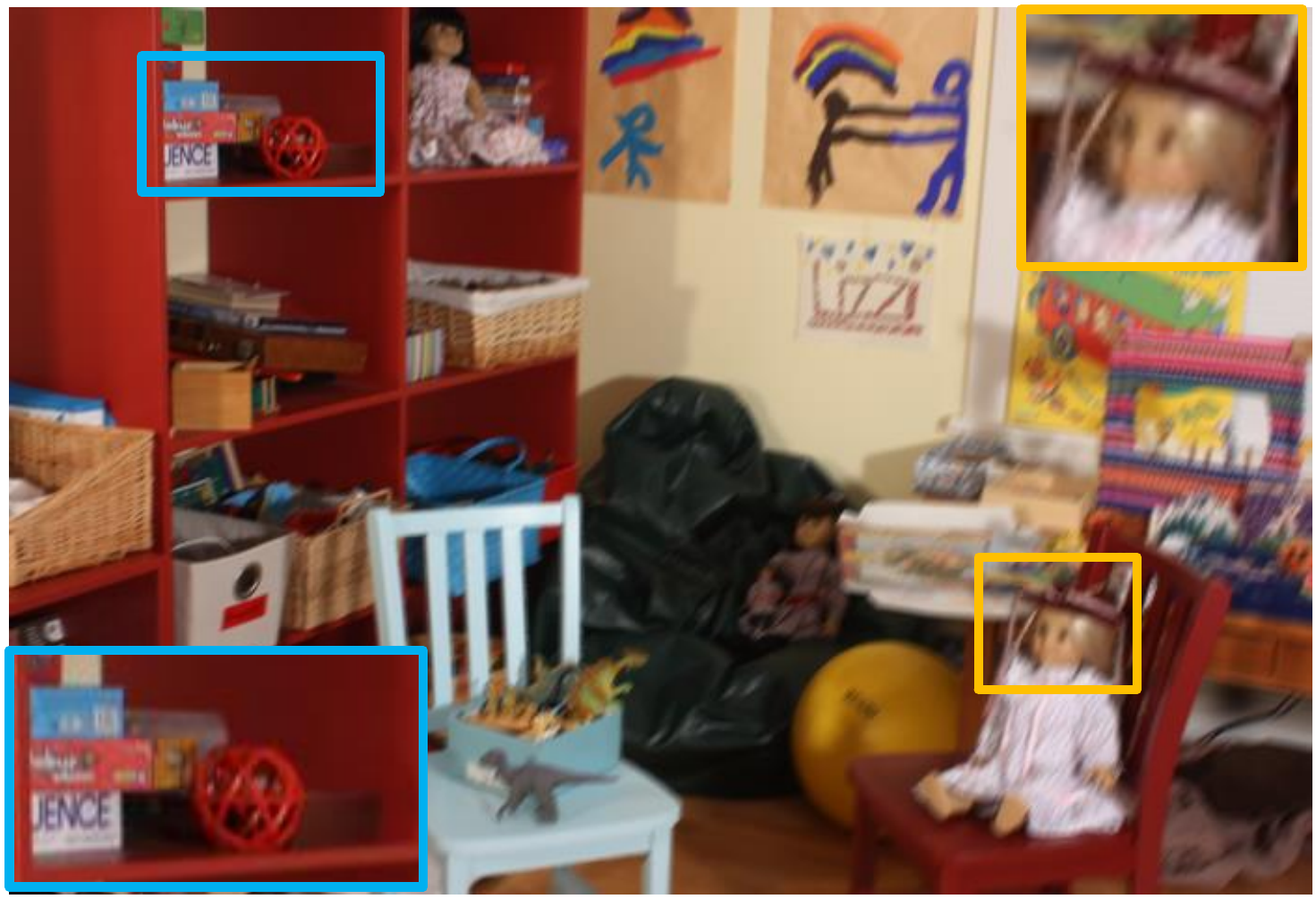}
&\includegraphics[width=0.23\textwidth]{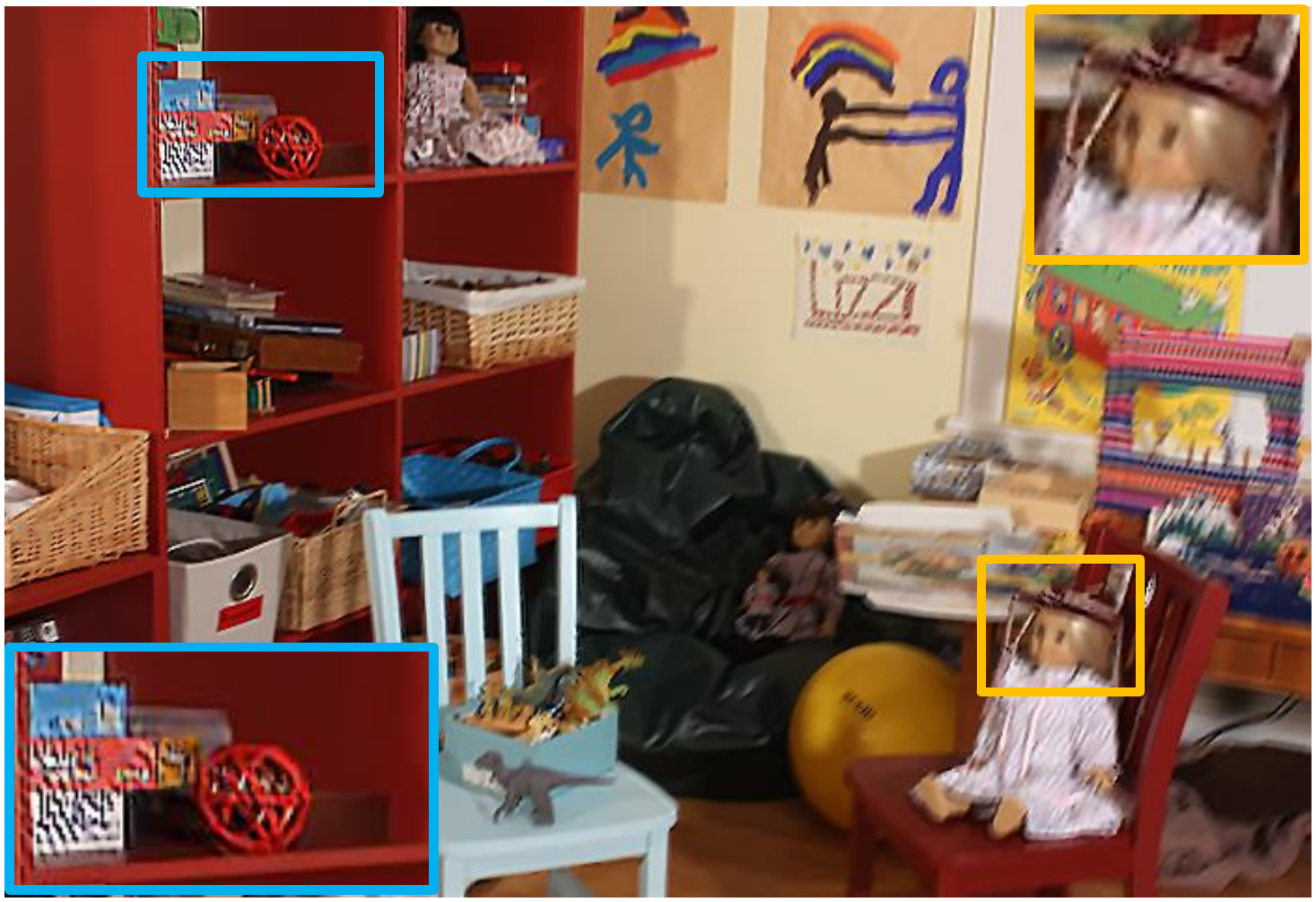}
&\includegraphics[width=0.23\textwidth]{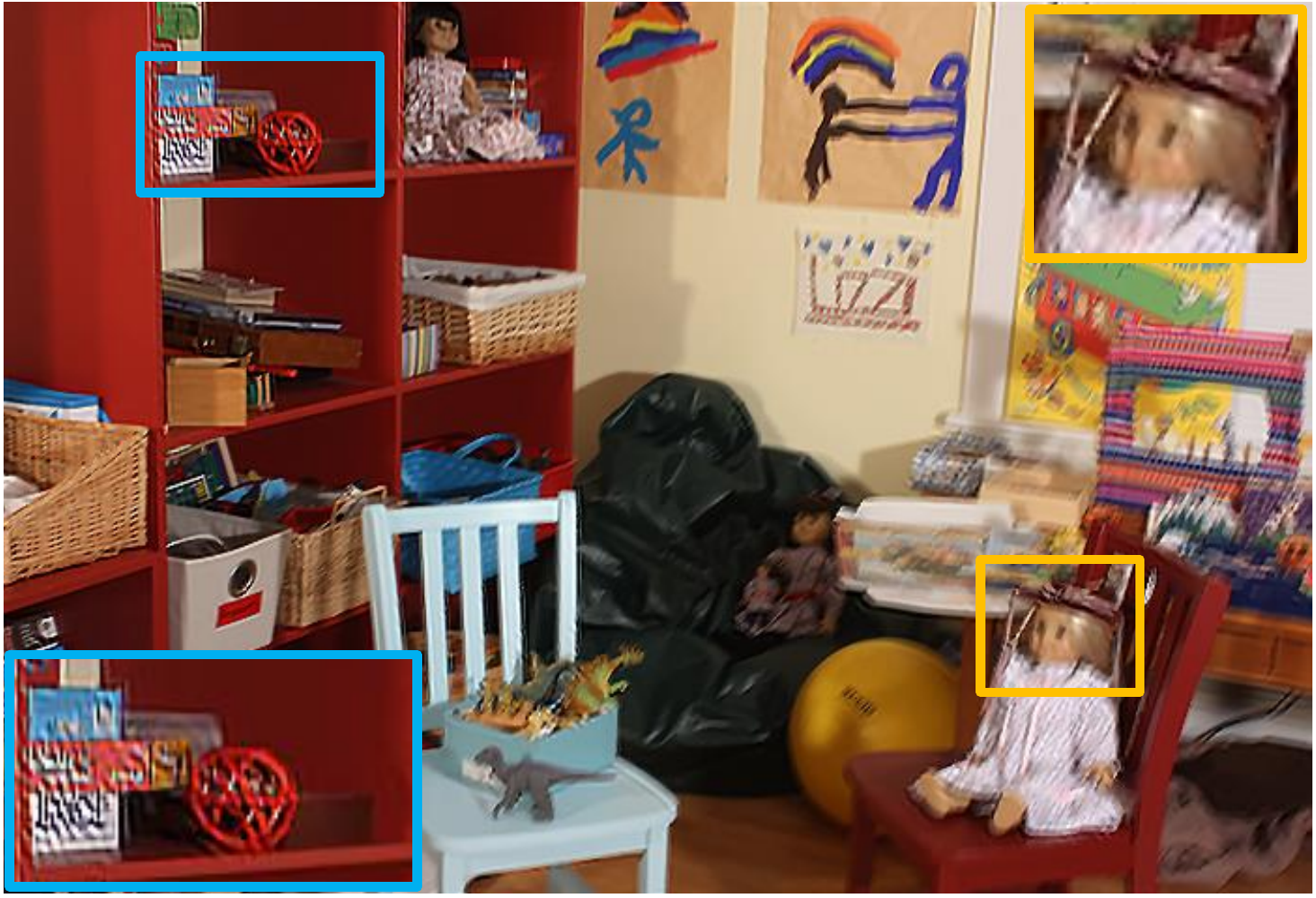}
&\includegraphics[width=0.23\textwidth]{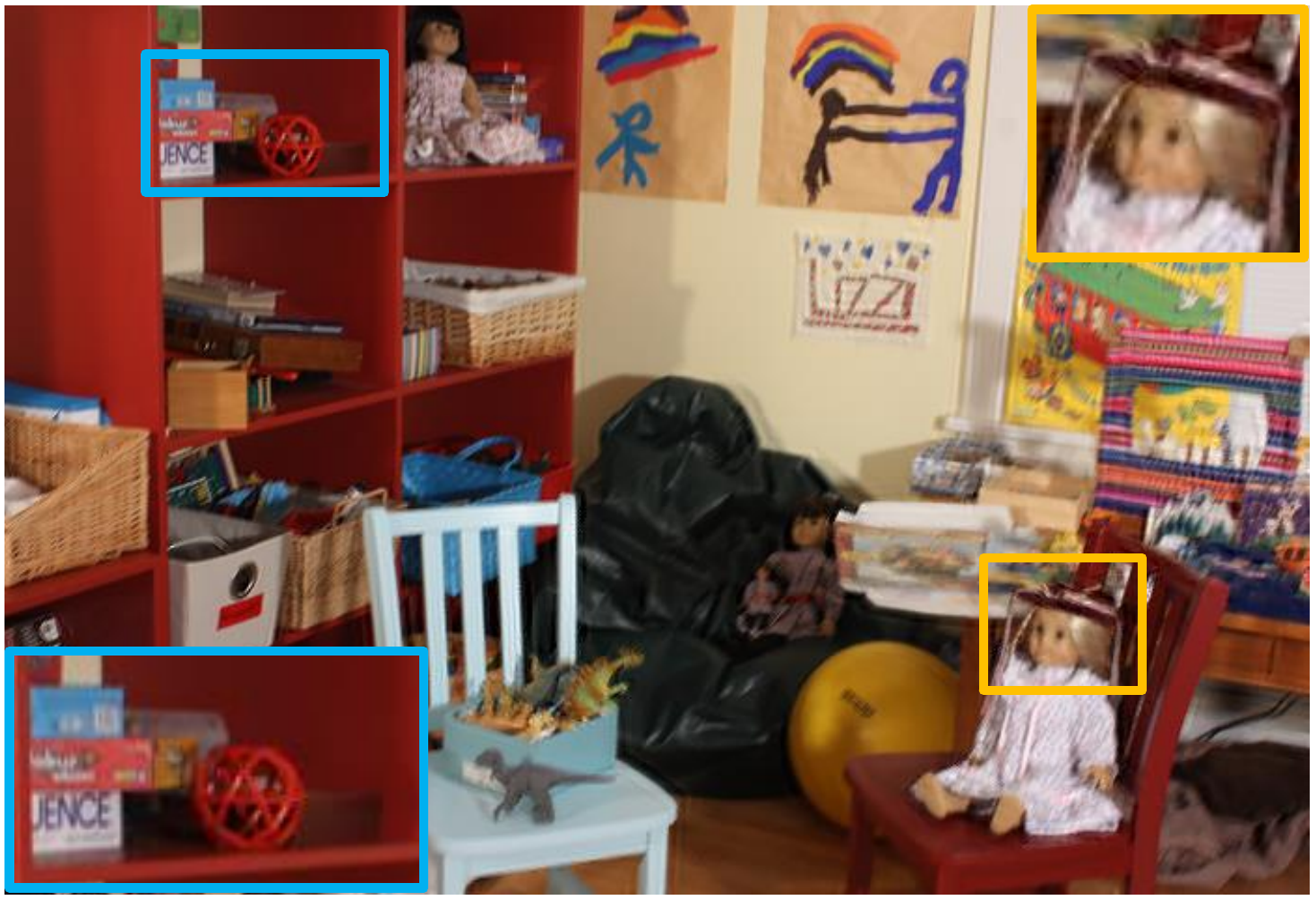}\\
\hspace{-0.35 cm}
\includegraphics[width=0.23\textwidth]{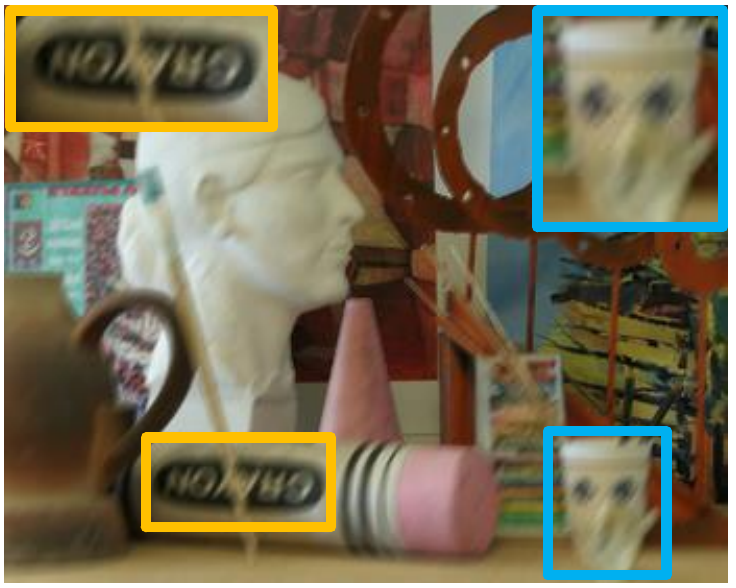}
&\includegraphics[width=0.23\textwidth]{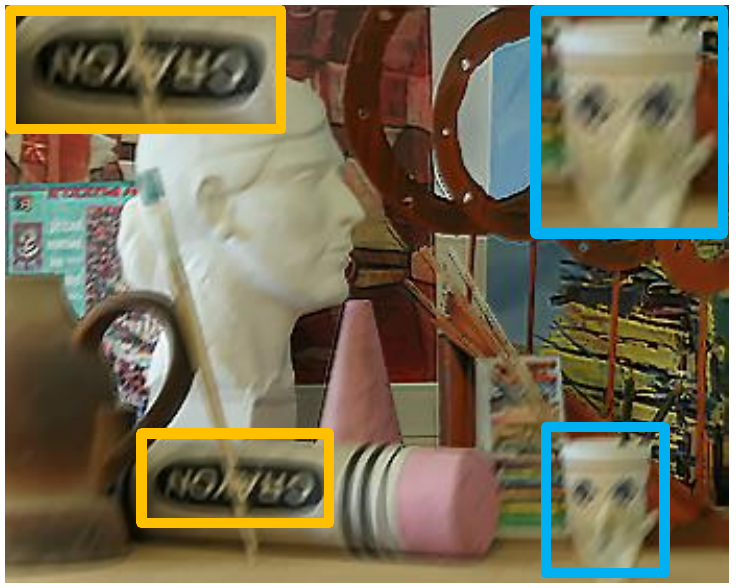}
&\includegraphics[width=0.23\textwidth]{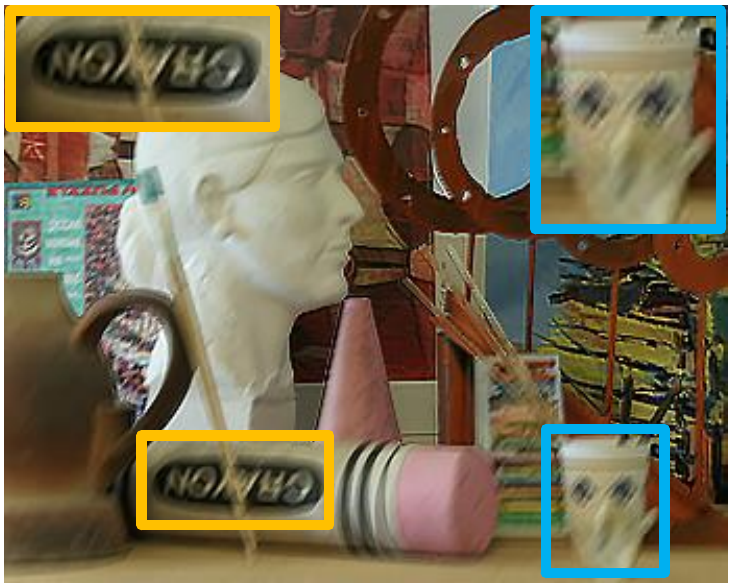}
&\includegraphics[width=0.23\textwidth]{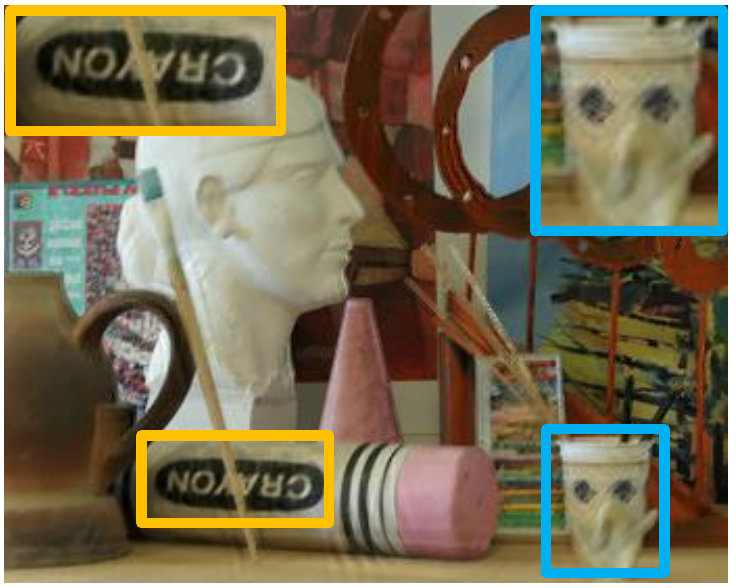}\\
\hspace{-0.35 cm}
(a) Input Blurry Image
& (b) Pan \cite{pan2017deblurring}
& (c) Yan \cite{yan2017image}
& (d) Ours\\
\end{tabular}
}
\end{center}
\caption{\label{fig:midd}
Example deblurring results on the Middlebury dataset. (a) Input blurry color images. (b) Deblurring results of \cite{pan2017deblurring}. (c) Deblurring results of \cite{yan2017image}. (d) Our deblurring results. (Best viewed on screen).  
}
\end{figure*}


\noindent{\bf{Synthetic Datasets.}} To the best of our knowledge, there are no realistic benchmark datasets that provide blurry images, their corresponding ground-truth depth maps,  and the latent clean images. We thus make use of the KITTI~\cite{geiger2013vision} and Middlebury dataset~\cite{scharstein2014high} to create synthetic datasets on realistic scenery. Since the camera shake always involves small rotation and translation, we thus sample the rotation angle for each image from a Gaussian distribution with the standard deviation $\sigma_a = 0.05~\mathrm{rad}$ and translation vector from a Gaussian distribution with $\sigma_t = 0.4\mathrm{m}$ for KITTI and $\sigma_t = 0.05\mathrm{m}$ for Middlebury.  The difference in the standard deviation is to match the different depth range in two datasets, which is $3\mathrm{m}$ for Middlebury and $40\mathrm{m}$ for KITTI dataset, respectively.

The blurry image is generated by averaging the~\emph{captured clean images} at $N=20$ uniformly distributed camera motion and locations within the exposure time $T=0.23$ (see Eq.~\ref{eq:convBlurKernel} and Fig.~\ref{fig:pipeline} for details). In particular, the clean images are rendered based on the camera motion in 3D space. Note that the blurry image rendering process requires a dense depth map. Instead of filling in holes for the sparse raw depth map in KITTI, we adopt the unsupervised stereo matching approach~\cite{zhong2017self}, which ranks among the methods of top performance on KITTI dataset with a pre-trained model available, to estimate the dense disparity map referred to as oracle depth. We create our testing set using 200 images chosen from different image sequences in KITTI. We similarly generate the testing set with 14 images from Middlebury 2014 using the depth maps provided by the dataset.

\vspace{0.5mm}
\noindent{\bf{Real Dataset.}} We further evaluate our method on the TUM RGB-D dataset~\cite{sturm2012benchmark}, which includes both depth maps and real blurry images. The captured depth maps and color images are of size $640 \times 480$. The measurements from the depth sensors are imperfect, which are noisy and contaminated with large holes due to the reflective surfaces and distant objects in the scene. We thus pre-process the depth maps by filling in those holes using a traditional depth completion method~\cite{yang2014color}. We test our algorithm on 300 images chosen from the 'bear' and 'walkman' sequences. Since the TUM dataset does not include ground truth sharp images, we thus only provide qualitative comparison with the state-of-the-art blind deblurring approaches.

\vspace{0.5mm}
\noindent{\bf{Implementation Details.}}  We validate the parameters in our model on three reserved images for each dataset. We set $\mu_1 = -20$, $\mu_2 = \mu_3 = 0.2$, $\mu_4 =0.05$, $\sigma_{\mathrm{B}} = 0.01$, $\sigma_{\mathrm{D}} = 0.02$ for all of our experiments.  In order to give a better initialization for our method, we first apply a conventional blind de-convolution approach~\cite{krishnan2011blind} to estimate a uniform blur kernel of size $25 \times 25$ to provide a prior on our 6 DoF pose $\mathbf{p}$. Our experiments show that such initialization is more robust than initializing the algorithm randomly. We further implement our algorithm in the traditional coarse-to-fine manner to achieve fast convergence. In particular, the image pyramid is built with 11 levels and the scale factor is set as 0.9. The motion parameters and the latent image estimated from coarse resolution are propagated as initialization to the next pyramid level. Our framework is implemented using MATLAB with C++ wrappers. It takes around 5 minutes to process one image on a single i7 core running at 3.6 GHz.

\vspace{1.5mm}
\noindent{\bf{Baselines and Evaluation Metric.}} We compare our approach with the state-of-the-art blind deblurring methods, such as ~\cite{yan2017image}, ~\cite{pan2017deblurring} and \cite{hu2014joint}, which handle spatially variant blur from a single image.
We further compare with a video method ~\cite{hyun2015generalized} and two learning based methods \cite{gong2017motion,Nah_2017_CVPR}which can handle non-uniform blur on the TUM dataset.

We report the PSNR and SSIM on our deblurred images. Instead of directly evaluating the rotation and translation estimation, we report the optical flow errors which are introduced by the errors in the camera motion estimation. In particular, the error metric is computed by counting the number of pixels which have errors more than 3 pixels and $5\%$ of its ground-truth.

\begin{figure*}[!ht]
\begin{center}
\begin{tabular}{cccc}
\includegraphics[width=0.280\textwidth]{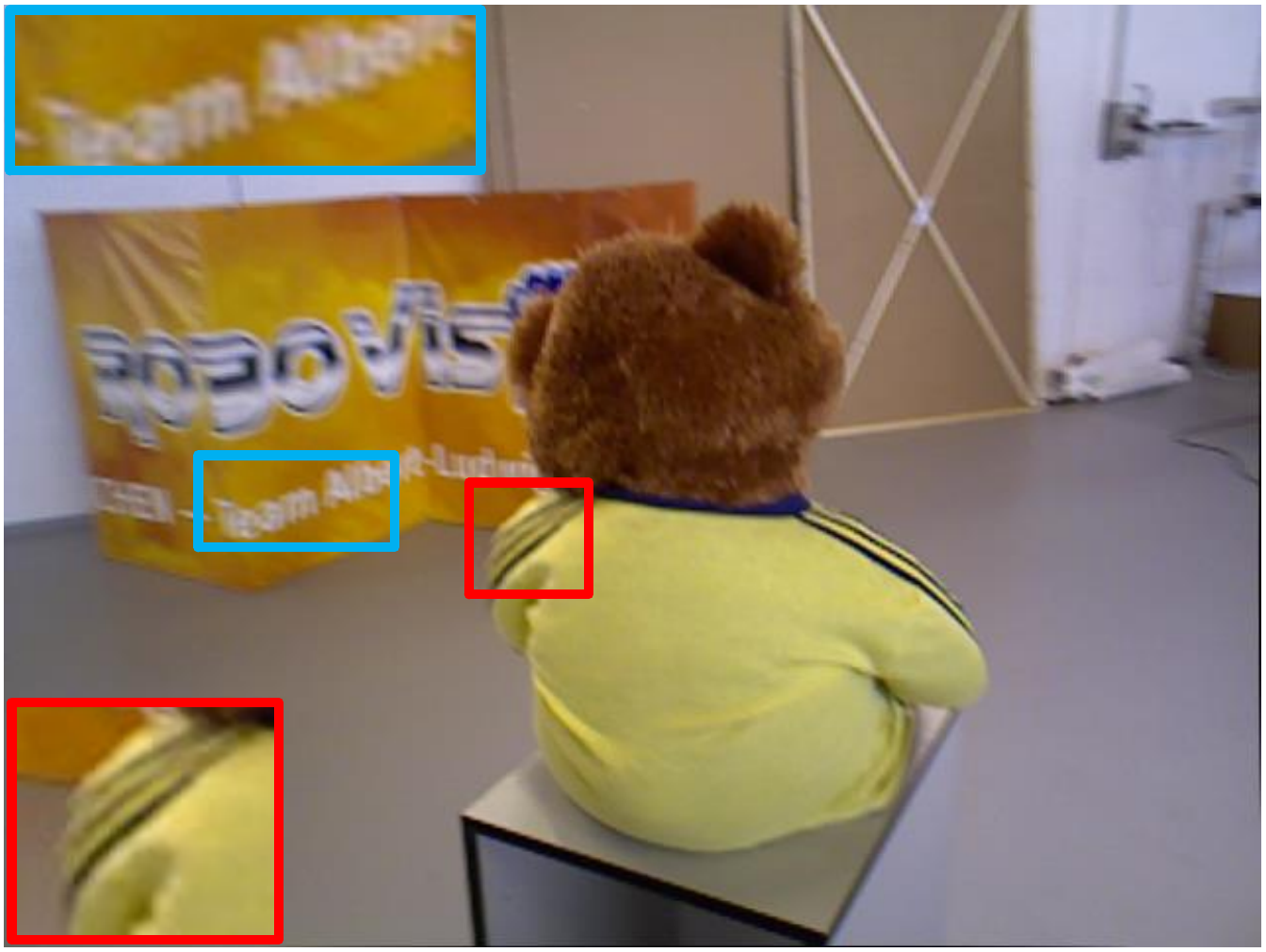}
&\includegraphics[width=0.280\textwidth]{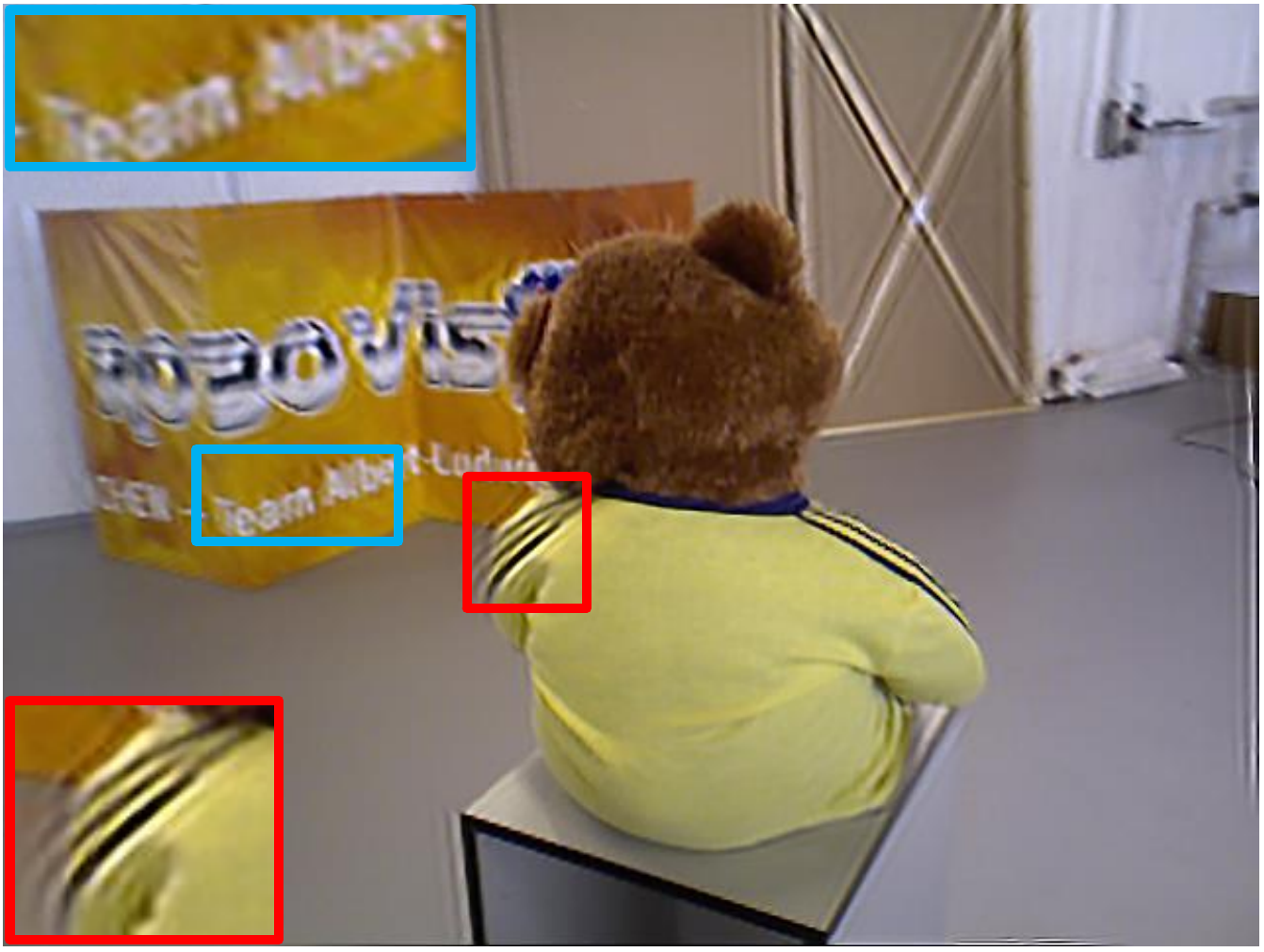}
&\includegraphics[width=0.280\textwidth]{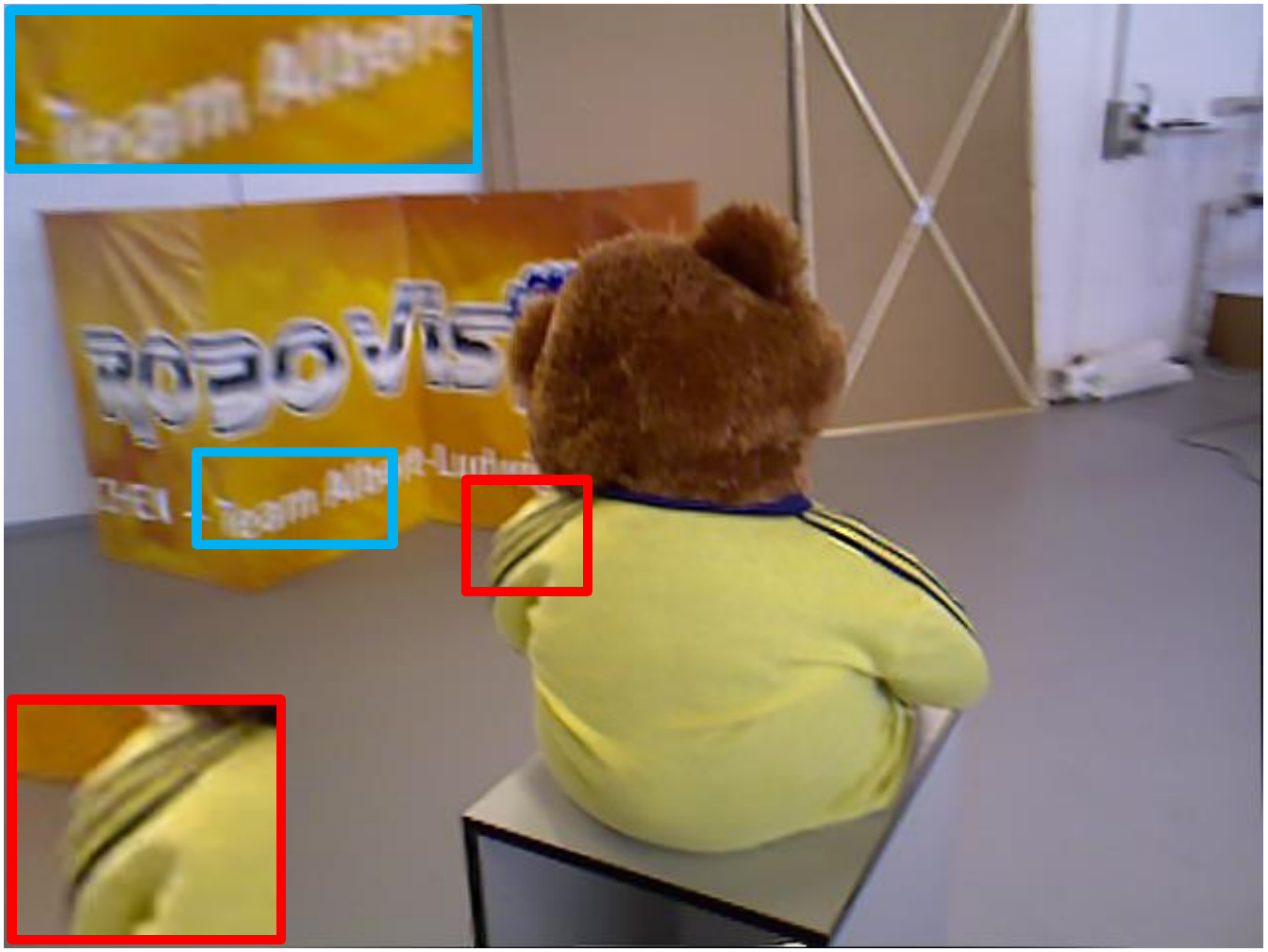}\\
(a) Blurry Image
& (b) Kim \cite{hyun2015generalized}
& (c) Gong \cite{gong2017motion} \\
\includegraphics[width=0.280\textwidth]{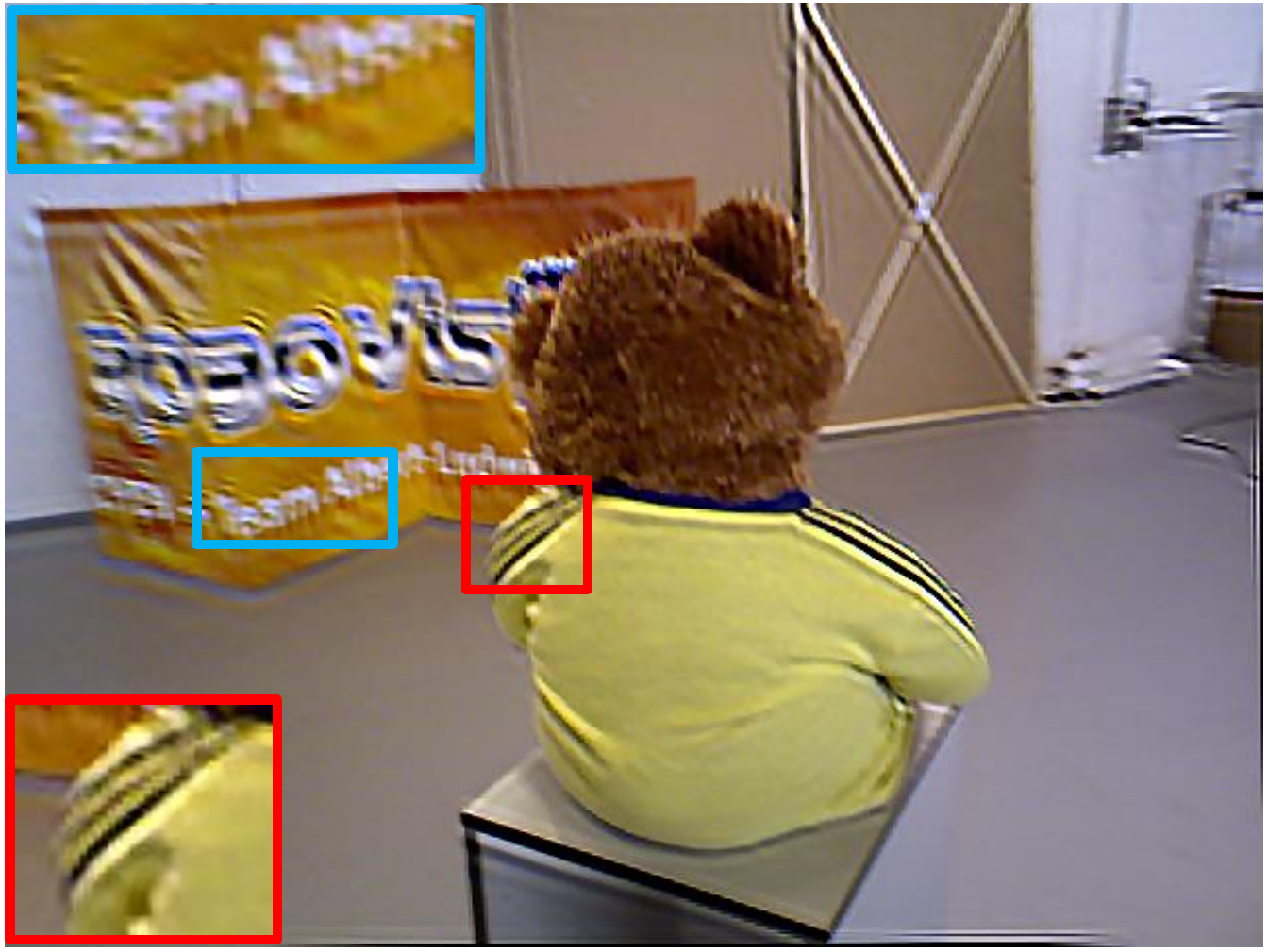}
&\includegraphics[width=0.280\textwidth]{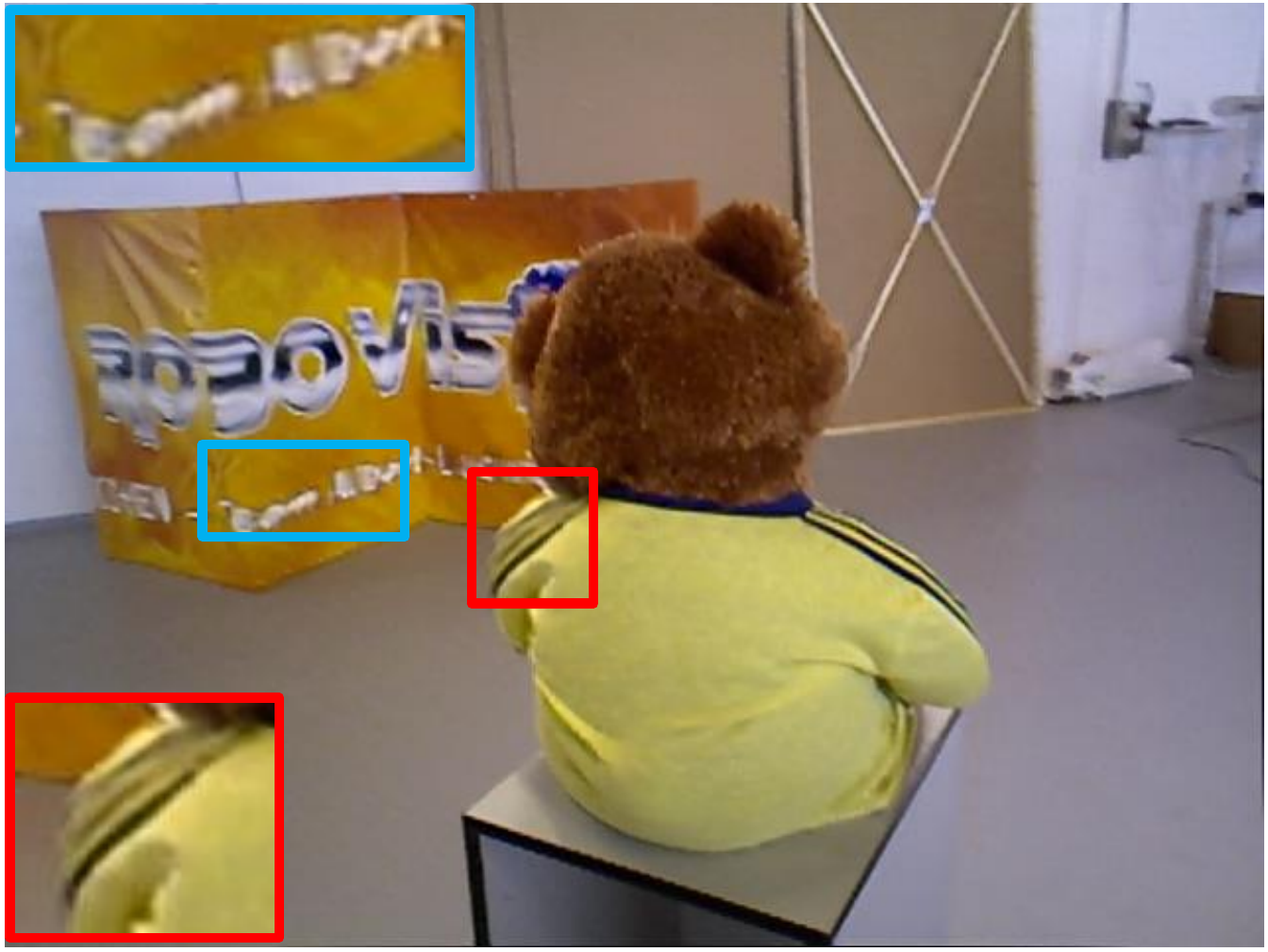}
&\includegraphics[width=0.280\textwidth]{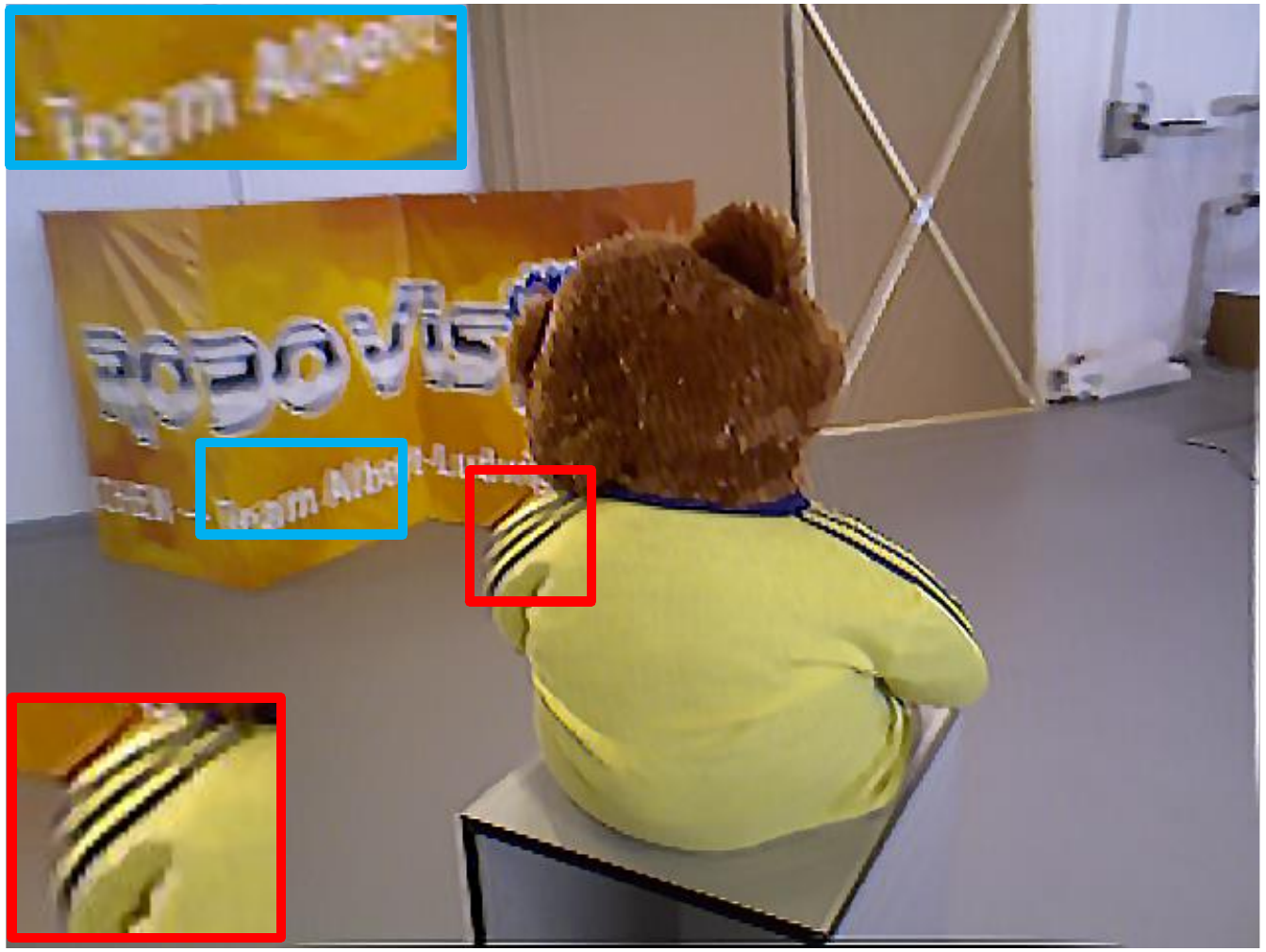}\\
(d) Hu \cite{hu2014joint}
& (e) Nah \cite{Nah_2017_CVPR}
& (f) Ours \\
\end{tabular}
\end{center}
\caption{\label{fig:TUM}
Comparison with the state-of-the-art non-uniform deblurring methods using real blurry image from the TUM dataset. The depth is from the Kinect sensor.
(a) Blurry image. (b) Video based deblurring result \cite{hyun2015generalized}. (c) Learning based result \cite{gong2017motion}. (d) Single image based deblurring result \cite{hu2014joint}, which also considers depth in their formulation. (e) Learning based result \cite{Nah_2017_CVPR}. (f) Our deblurring result. 
}
\end{figure*}

\subsection{Experimental Results}
\begin{figure*}
\begin{center}
\resizebox{\textwidth}{!}{
\begin{tabular}{cc}
\includegraphics[width=0.47\textwidth]{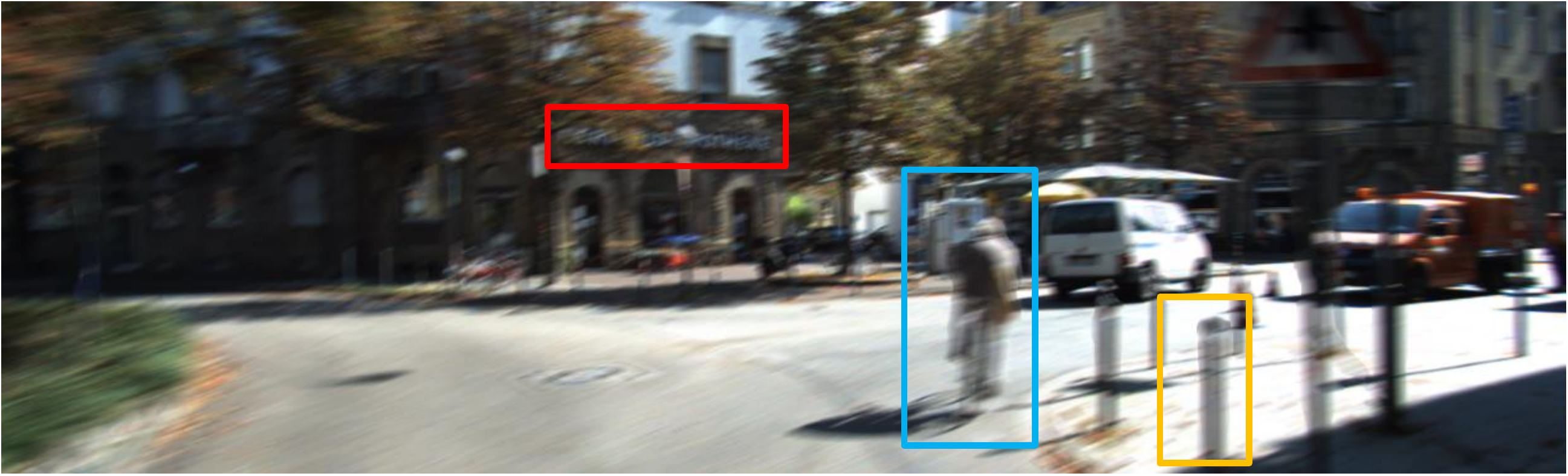}
&\includegraphics[width=0.47\textwidth]{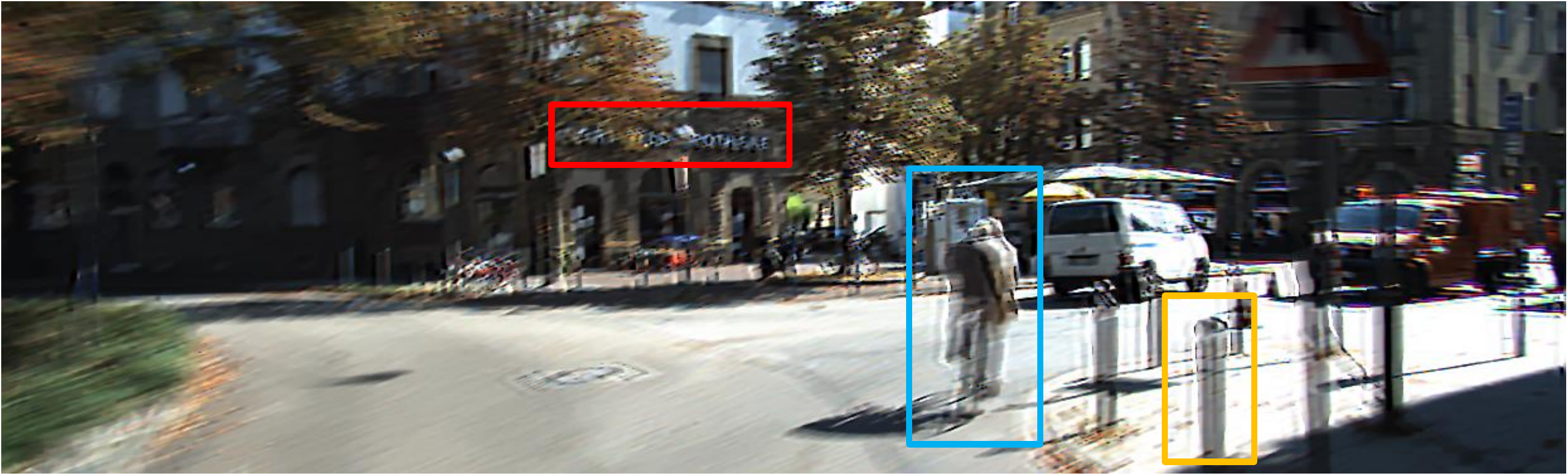}\\
\includegraphics[width=0.47\textwidth]{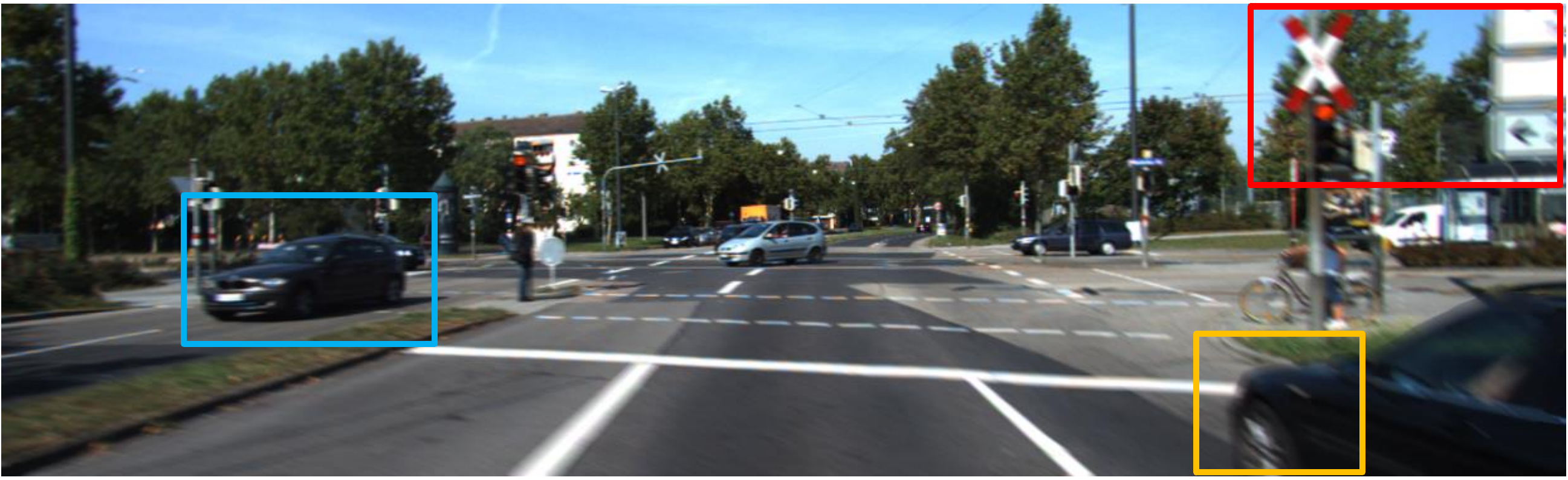}
&\includegraphics[width=0.47\textwidth]{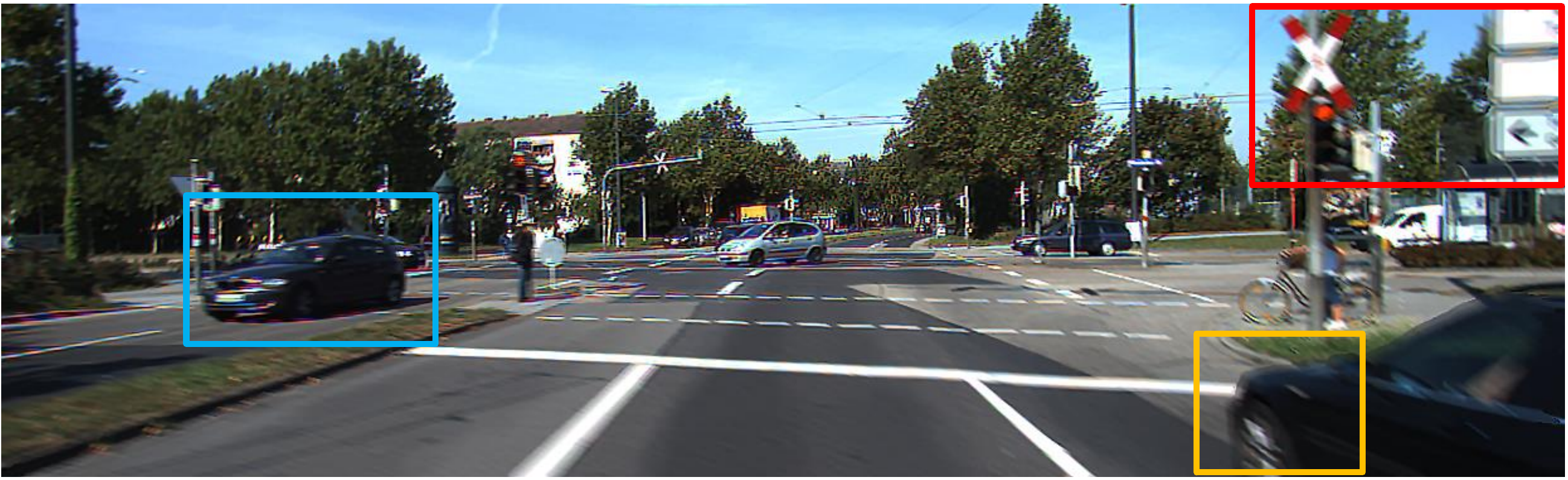}\\
\includegraphics[width=0.47\textwidth]{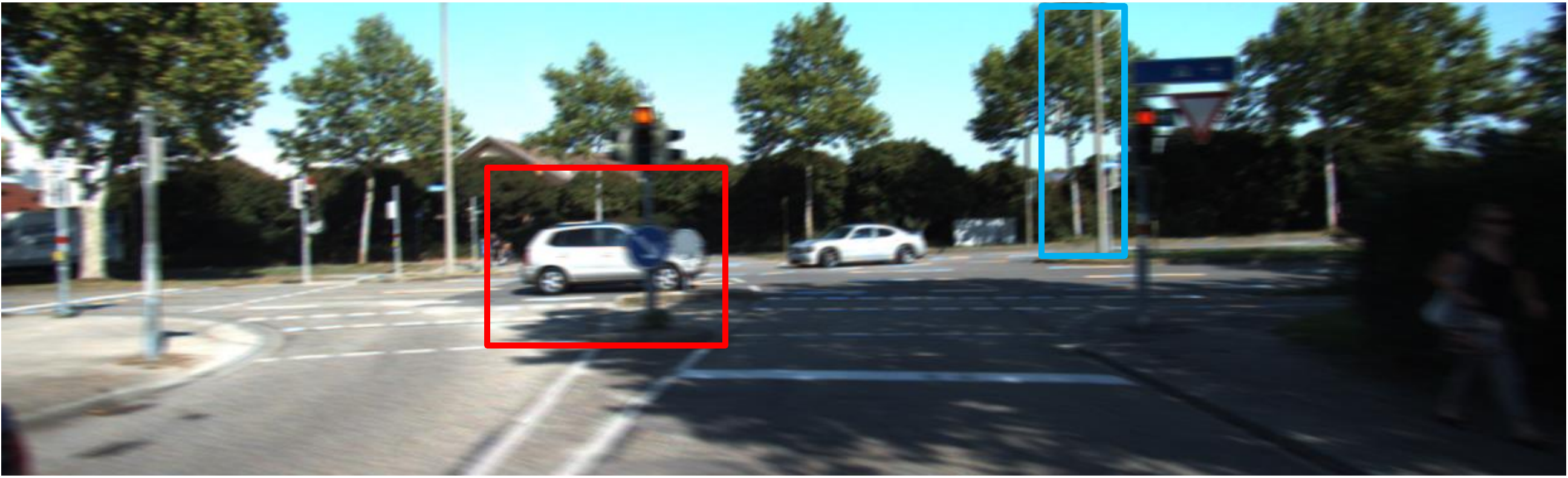}
&\includegraphics[width=0.47\textwidth]{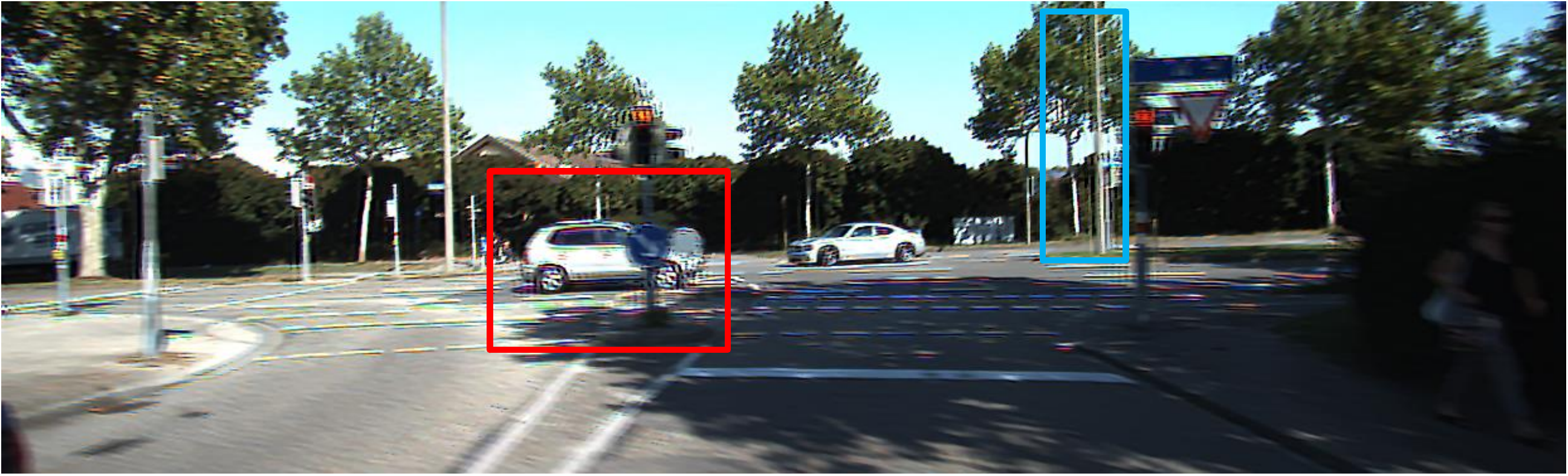}\\
(a) Input Blurry Image
& (b) Pan \cite{pan2017deblurring}\\
\includegraphics[width=0.47\textwidth]{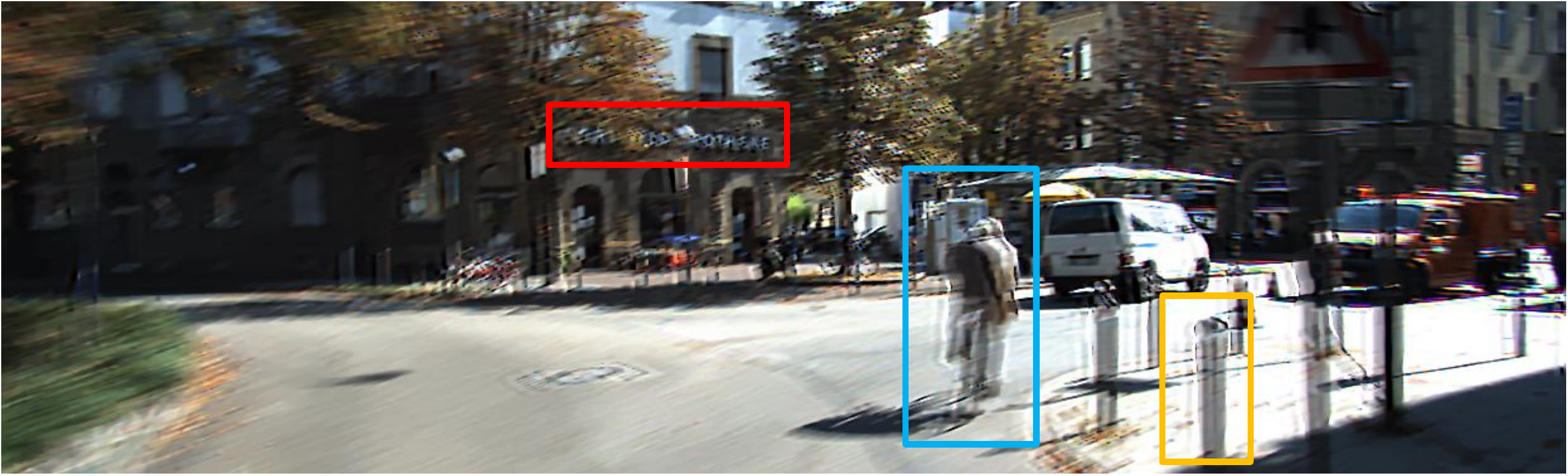}
&\includegraphics[width=0.47\textwidth]{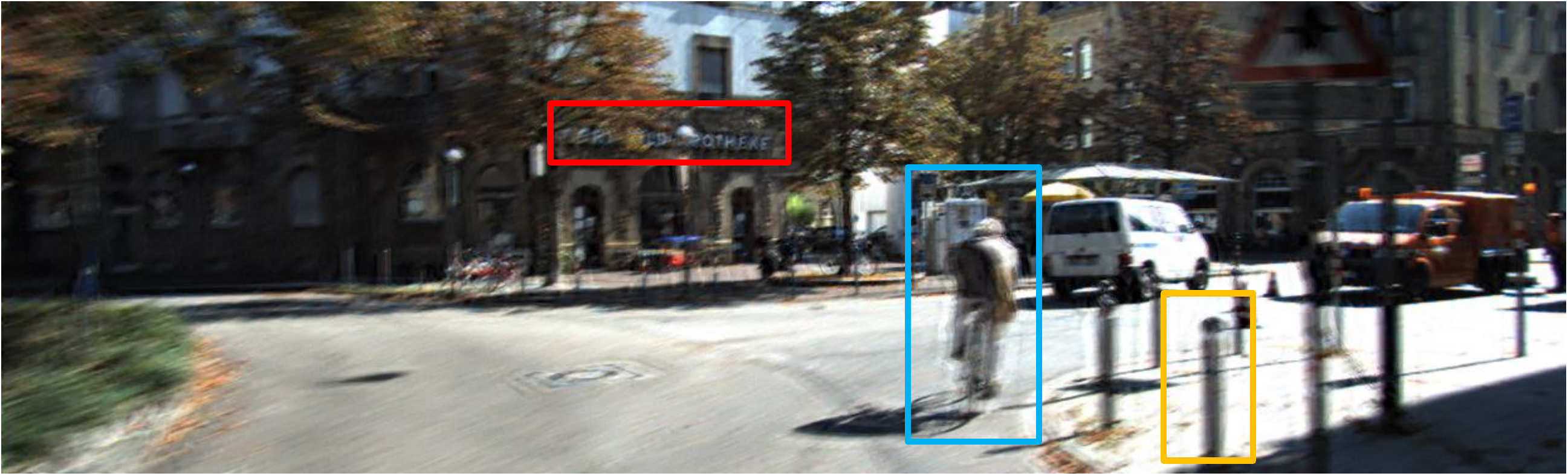}\\
\includegraphics[width=0.47\textwidth]{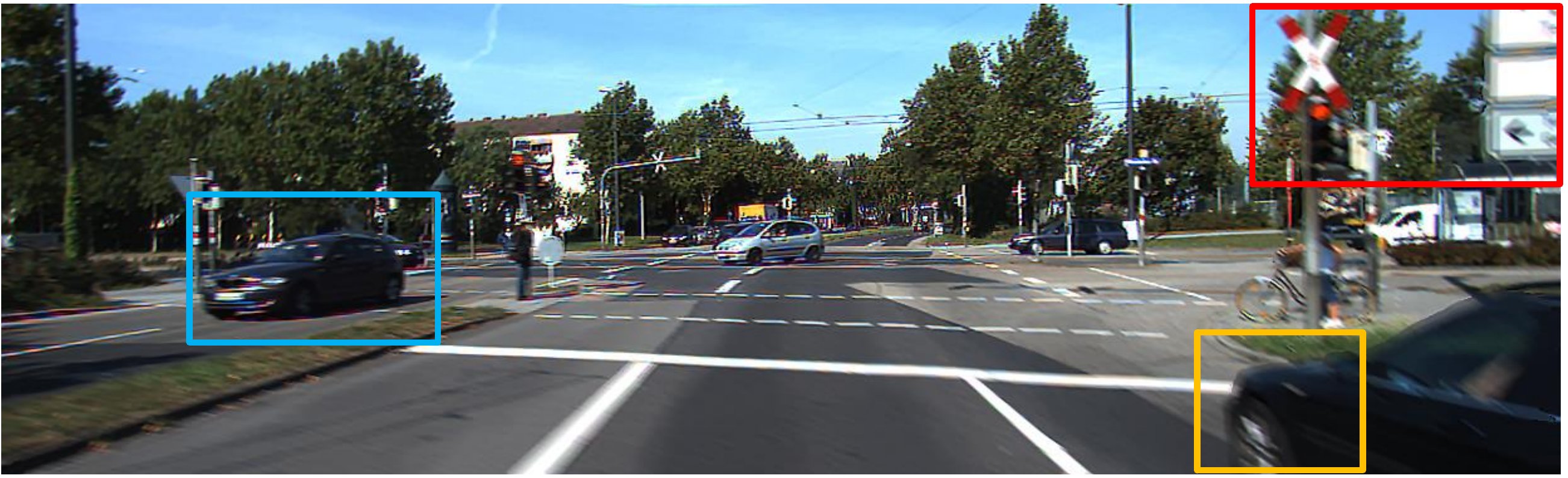}
&\includegraphics[width=0.47\textwidth]{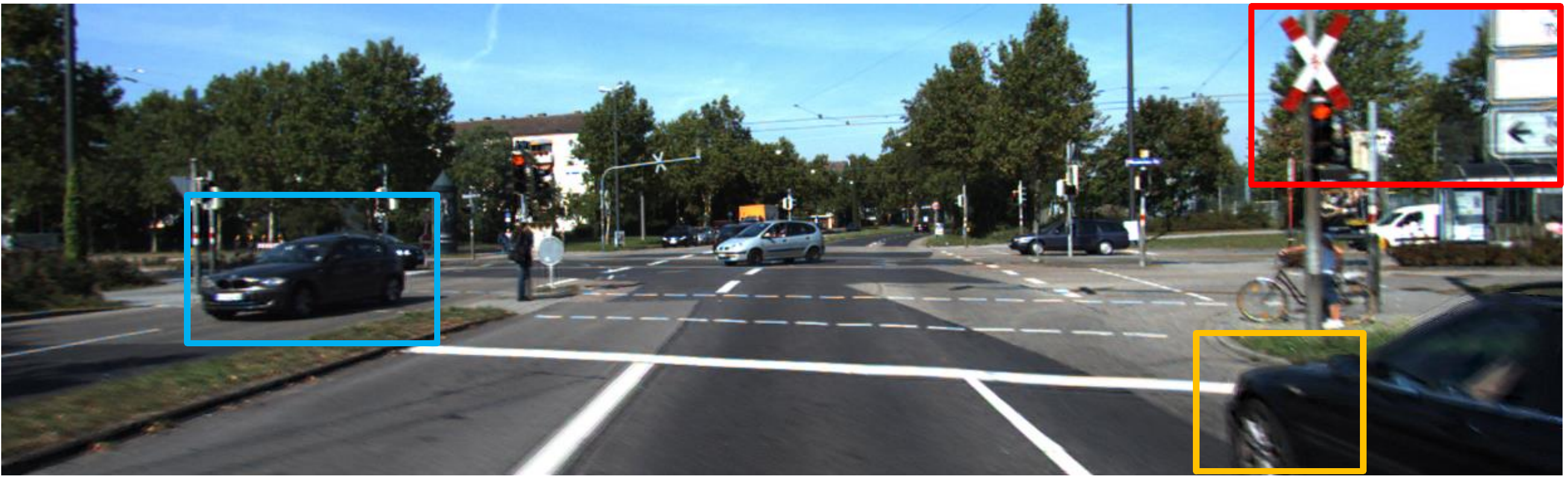}\\
\includegraphics[width=0.47\textwidth]{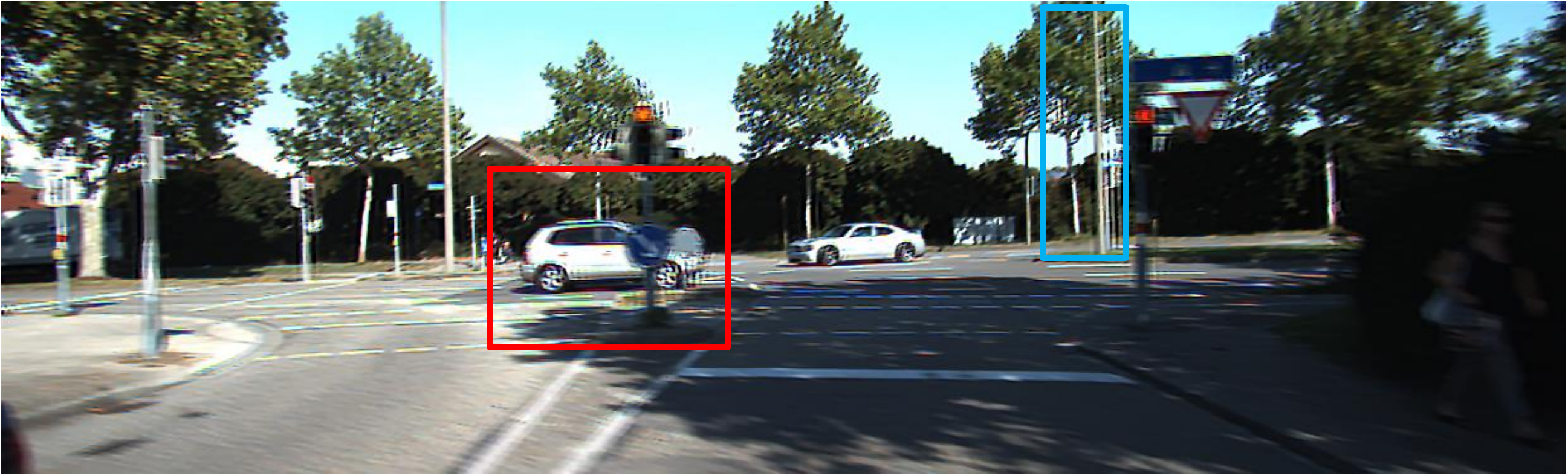}
&\includegraphics[width=0.47\textwidth]{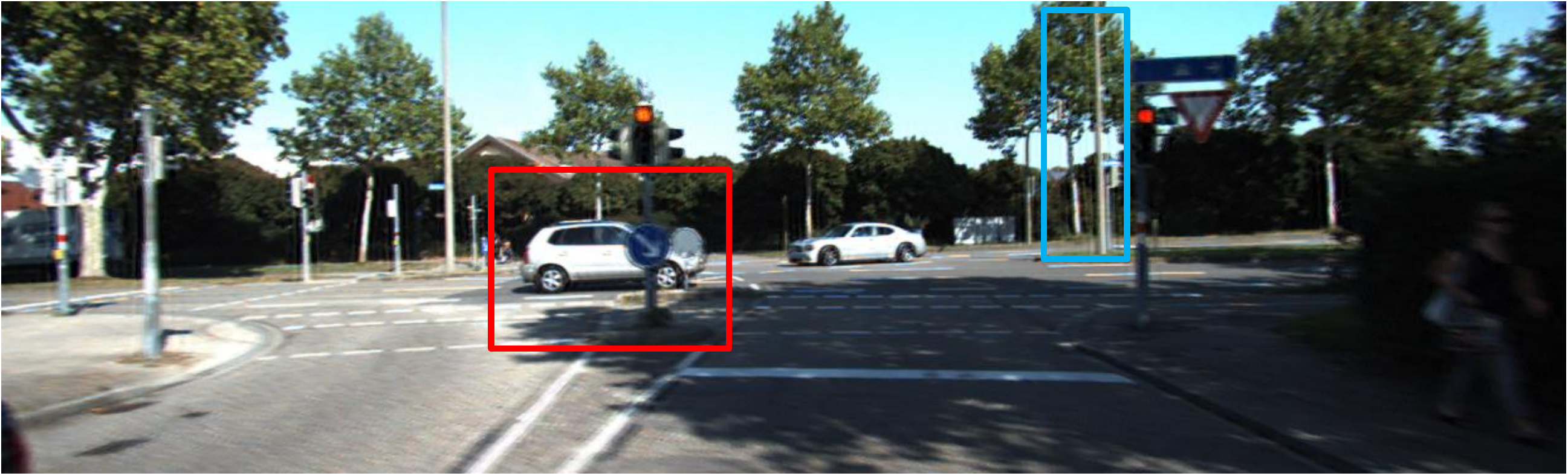}\\
 (c) Yan \cite{yan2017image}
& (d) Ours\\
\includegraphics[width=0.47\textwidth]{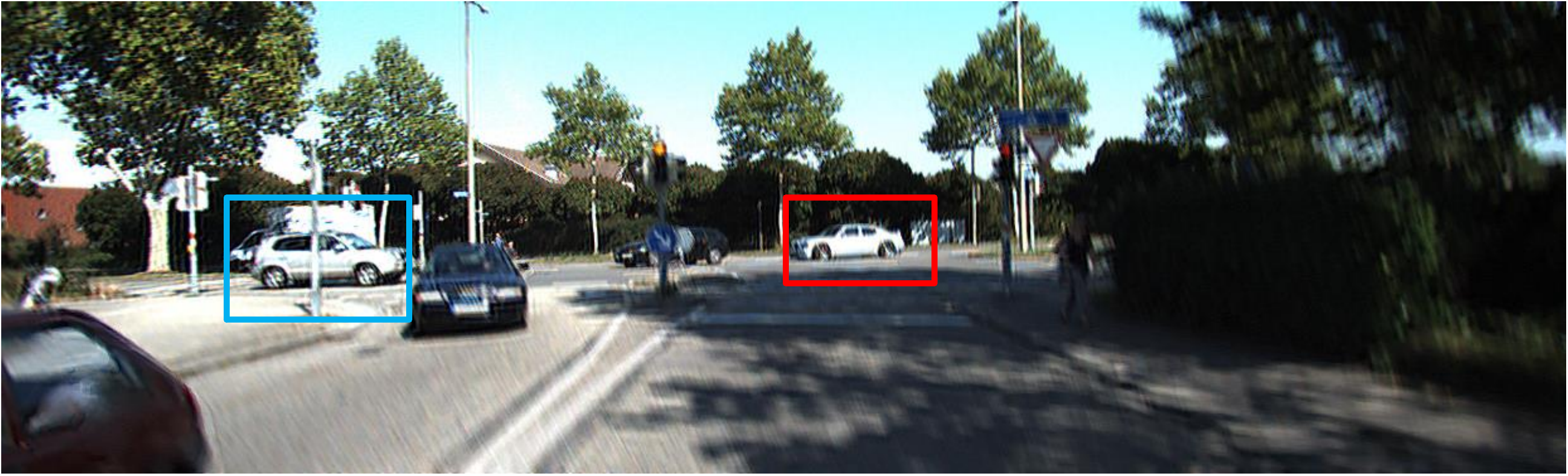}
&\includegraphics[width=0.47\textwidth]{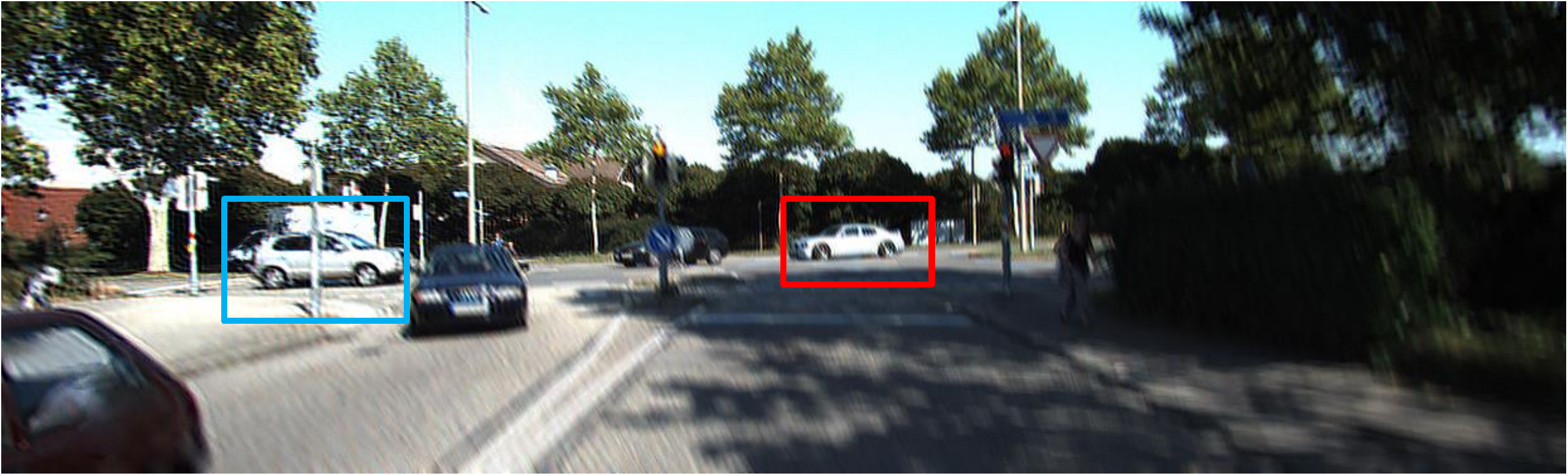}\\
{\bf (e)Input with oracle depth from \cite{zhong2017self} } & {\bf (f) Input with learned depth from~\cite{godard2017unsupervised}}
\end{tabular}
}
\end{center}
\caption{\label{fig:KITTI}
Example deblurring results on the KITTI dataset. (a) Input blurry color images. (b) Deblurring results of \cite{pan2017deblurring}. (c) Deblurring results of \cite{yan2017image}. (d) Our deblurring results with learned depth map as input. In order to compare the results with respect to different input depth map, {\bf (e) and (f) show our deblurring results with oracle depth map and learned depth map as inputs, respectively.} Compared with the two state-of-the-art deblurring methods, our method achieves the best performance (Best viewed on screen). }
\end{figure*}

For the above datasets we used, the depth is from stereo matching, depth sensor and learned by neural network. In Table~\ref{all}, we compare our approach with the state-of-the-art single image deblurring methods,~\cite{yan2017image} and~\cite{pan2017deblurring}, for spatially-variant blurs on Middlebury, KITTI, and TUM dataset, based on the PSNR, SSIM and Flow Error metric. Note that experiments on Middlebury and TUM used depth with high accuracy (provided by the dataset) as input for deblurring. In order to evaluate the robustness of our approach w.r.t. the depth quality, we adopted the most recent unsupervised monocular depth estimation method~\cite{godard2017unsupervised} to learn the depth maps for KITTI dataset as input to remove the blurs generated based on oracle depth from stereo matching approach~\cite{zhong2017self}. We further provide an example for visual comparison in Fig.~\ref{fig:KITTI}(e),(f) to show the difference of the deblurring results from the oracle depth and the learned depth, respectively. More comparisons will be included in the supplementary material.~ Note that our approach outperforms all the baseline approaches which do not reason about the camera motion, by a large margin. This evidences the importance of our joint camera motion estimation and image deblurring framework. We further compare our approach with the image deblurring approach from monocular video sequence~\cite{sturm2012benchmark}. This again shows the importance of including depth information and performing 6 DoF camera motion estimation for blind deblurring.


The qualitative comparisons on the three datasets are shown in Fig.~\ref{fig:midd},~\ref{fig:KITTI}, and~\ref{fig:TUM}, respectively. The qualitative results show that our approach can recover more sharp details than other competing approaches, which are highlighted in the reported results. Note that our deblurring results can recover the color images more faithfully than the baselines. It further evidences the quantitative improvements as shown in Table~\ref{all}. Last but not least, as we have recovered the 6 DoF camera motion, we can generate a sharp video sequence correspondingly as illustrated in Fig.~\ref{fig:fig1}, where each novel frame is generated by warping the latent clean image with the corresponding camera motion estimation and estimated/measured depth maps.


\section{Conclusions}
In this paper, we have presented a joint optimization framework to estimate the 6 DoF camera motion and deblur the image from a single blurry image. To alleviate the difficulties, we exploit the availability of depth maps (either from noisy measurements or learned through a deep neural network) and a small motion model for the camera. Under our formulation, the solution of one sub-task benefits the solution of the other sub-task. Extensive experiments on both synthetic and real image datasets demonstrate the superiority of our framework over very recent state-of-the-art blind image deblurring methods such as dark channel prior \cite{pan2017deblurring} and extreme channel prior \cite{yan2017image}). In the future, we plan to exploit more general parametric camera trajectories to further improve the performance in real world challenging scenarios. 

\section*{Acknowledgements}
This work was supported in part by Natural Science Foundation of China grants (61871325, 61420106007, 61671387, 61603303) and the Australian Research Council (ARC) grants (DE140100180, DE180100628).%


{\small
\bibliographystyle{ieee}
\bibliography{blurwithdepthbib}
}

\end{document}